\documentclass[lettersize,journal,twocolumn]{IEEEtran}
\usepackage{amsmath,amsfonts}
\usepackage{algorithmic}
\usepackage{algorithm}
\usepackage{array}
\usepackage[caption=false,font=normalsize,labelfont=sf,textfont=sf]{subfig}
\usepackage{textcomp}
\usepackage{stfloats}
\usepackage{url}
\usepackage{verbatim}
\usepackage{graphicx}
\usepackage{cite}
\usepackage{hyperref}
\usepackage{multirow}
\usepackage{tabularx}
\usepackage{cuted}
\usepackage{float}
\usepackage{booktabs}
\usepackage{pifont}
\usepackage{bbding}
\usepackage[table,dvipsnames]{xcolor}
\newcommand{\best}{\cellcolor{red!50}}
\newcommand{\sbest}{\cellcolor{orange!80}}
\newcommand{\tbest}{\cellcolor{yellow!80}}
\hypersetup{
colorlinks=true,
linkcolor=blue,
citecolor=blue
}
\usepackage{xcolor}
\begin{document}

\title{VDNeRF: Vision-only Dynamic  \\ Neural Radiance Field for Urban Scenes}

\author{Zhengyu Zou, Jingfeng Li, Hao Li, Xiaolei Hou, Jinwen Hu, Jingkun Chen,\\ Lechao Cheng\textsuperscript{\Envelope}, Dingwen Zhang\textsuperscript{\Envelope}
\thanks{This work was supported by National Natural Science Foundation of China (No. 62293543, 62322605, and 62472139). (Corresponding authors: Lechao Cheng and Dingwen Zhang.)}
\thanks{Zhengyu Zou, Jingfeng Li, Hao Li, Xiaolei Hou, Jinwen Hu, Dingwen Zhang are with Northwestern Polytechnical University, Xi'an, China (email:).}
\thanks{Jingkun Chen is with the Department of Engineering Science, University of Oxford, Oxford, OX3 7DQ, United Kingdom (email: chenjingkun@outlook.com) }
\thanks{Lechao Cheng is with Hefei University Of Technology, Hefei, China (email: chenglc@hfut.edu.cn).}
\thanks{Manuscript received November, 2025;}}
\markboth{CAAI Transactions on Intelligence Technology,~November~2025}%
{\shortauthors: \shorttitle}
\maketitle
\begin{strip}
\begin{minipage}{\textwidth}\centering
\vspace{-70pt}
\begin{figure}[H]
\centering
    \hsize=\textwidth
    \includegraphics[width= 0.85\textwidth]{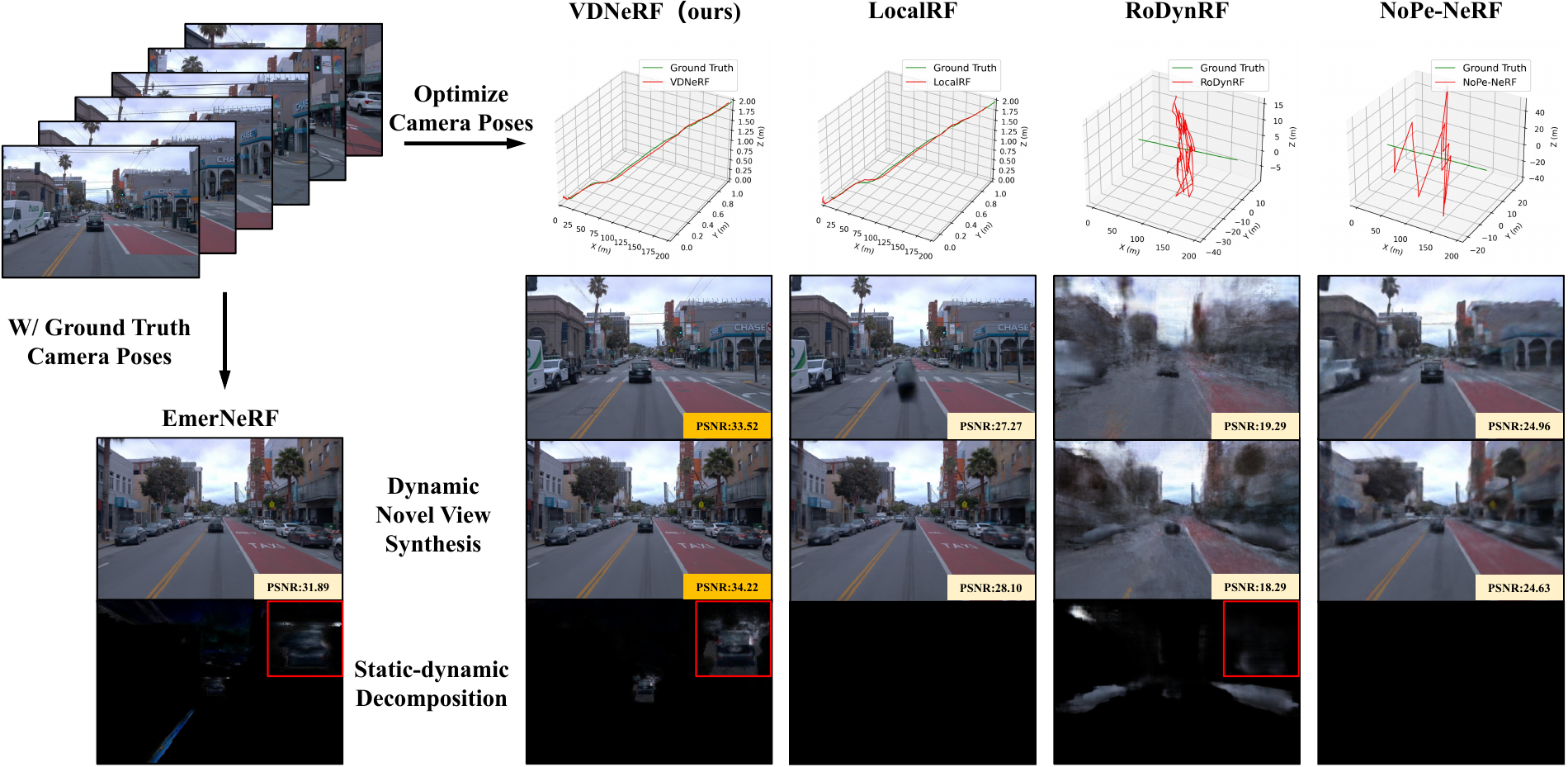}
    \caption{\textbf{Novel view synthesis for dynamic urban scenes.} Given only a set of images without known camera poses, VDNeRF can recover the camera trajectory and reconstruct the spatiotemporal scene. Compared with existing methods, VDNeRF demonstrates more robust camera pose estimation, higher-quality dynamic novel view synthesis, and more precise static-dynamic decomposition. We zoom in on the dynamic objects within the red box in the upper right corner to provide a more detailed comparison.}
    \label{fig:teaser}
\end{figure}
\end{minipage}
\end{strip}

\begin{abstract}
Neural Radiance Fields (NeRFs) implicitly model continuous three-dimensional scenes using a set of images with known camera poses, enabling the rendering of photorealistic novel views. However, existing NeRF-based methods encounter challenges in applications such as autonomous driving and robotic perception, primarily due to the difficulty of capturing accurate camera poses and limitations in handling large-scale dynamic environments. To address these issues, we propose Vision-only Dynamic NeRF (VDNeRF), a method that accurately recovers camera trajectories and learns spatiotemporal representations for dynamic urban scenes without requiring additional camera pose information or expensive sensor data. VDNeRF employs two separate NeRF models to jointly reconstruct the scene. The static NeRF model optimizes camera poses and static background, while the dynamic NeRF model incorporates the 3D scene flow to ensure accurate and consistent reconstruction of dynamic objects. To address the ambiguity between camera motion and independent object motion, we design an effective and powerful training framework to achieve robust camera pose estimation and self-supervised decomposition of static and dynamic elements in a scene. Extensive evaluations on mainstream urban driving datasets demonstrate that VDNeRF surpasses state-of-the-art NeRF-based pose-free methods in both camera pose estimation and dynamic novel view synthesis.
\end{abstract}

\begin{IEEEkeywords}
Neural Radiance Field (NeRF), computer vision, image-based localization, dynamic novel view synthesis, scene representation.
\end{IEEEkeywords}

\section{Introduction}
\IEEEPARstart{D}{ynamic} urban scene reconstruction using Neural Radiance Field (NeRF)~\cite{mildenhall2021nerf} has become a pivotal research area due to its ability to synthesize photorealistic images from arbitrary viewpoints. This capability is essential for applications such as robotic perception~\cite{nguyen2024mcd}, augmented reality (AR)~\cite{carmigniani2011augmented}, virtual reality (VR)~\cite{burdea2003virtual}, scene understanding~\cite{li2024gp, zhang2025unsupervised}, and autonomous driving~\cite{levinson2011towards}. However, traditional NeRF methods typically require precise camera pose information for each image, which is often challenging to obtain accurately in real-world scenarios~\cite{jiao2024instance}.

To estimate camera poses from images, Structure from Motion (SfM)~\cite{schonberger2016structure} techniques are commonly employed. Unfortunately, these methods can be time-consuming and may lack robustness, particularly in urban environments where dynamic elements like vehicles and pedestrians introduce variability. This variability can lead to multi-view inconsistencies across input images, making SfM less effective.

To eliminate the dependency on SfM, methods such as Nope-NeRF~\cite{bian2023nope} and LocalRF~\cite{meuleman2023progressively} have been developed to jointly optimize both camera poses and NeRF parameters by parameterizing camera poses. However, these approaches assume strictly static scenes, an assumption that is often violated in dynamic urban environments. The presence of moving objects introduces challenges in distinguishing between camera-induced motion and object-induced motion, complicating the optimization process and limiting the applicability of these methods in real-world urban scenarios.

RoDynRF~\cite{liu2023robust} extends NeRF-based pose-free reconstruction to dynamic scenes by employing object segmentation to generate 2D motion masks, thereby decoupling static and dynamic components. It utilizes a static radiance field to estimate camera poses and reconstruct the static background, while a dynamic radiance field, combined with a deformation field, reconstructs dynamic objects. However, RoDynRF faces limitations in effectively representing complex urban scenes, often resulting in visual artifacts and low-resolution outputs. Moreover, its reliance on precise 2D motion masks, which can be unreliable in urban settings, may lead to incorrect identification of stationary objects as dynamic, resulting in inaccurate scene representations.

Beyond these limitations, two key challenges must be addressed to achieve NeRF-based vision-only urban scene reconstruction and perception: \textbf{(I) Robust and Accurate Camera Pose Estimation:} In long and narrow urban scenes with dynamic objects, achieving precise camera pose estimation is crucial for modeling object motion and representing spatiotemporal scenes. However, existing methods often fall into local minima in such complex settings, leading to failures in recovering extensive camera trajectories. \textbf{(II) Self-Supervised Decomposition of Static and Dynamic Elements:} Effectively separating static backgrounds from dynamic foregrounds without precise annotations is vital for applications like identifying moving vehicles in autonomous driving. The joint optimization of camera poses and NeRF introduces challenges, as uncertainties in camera poses and the presence of dynamic objects can interfere with the static-dynamic decomposition process.

In this paper, we propose VD-NeRF (Vision-only Dynamic NeRF), a method that achieves novel view synthesis and spatiotemporal reconstruction for urban scenes using only a sequence of input images. VD-NeRF utilizes two separate NeRF models to represent dynamic scenes: \textbf{(a) Static NeRF Model}: Responsible for optimizing camera poses and reconstructing the static background; \textbf{(b) Dynamic NeRF Model}: Reconstructs dynamic objects by introducing temporal encoding and employs a flow field to predict 3D scene flow, ensuring consistency and accuracy in modeling dynamic objects.

Additionally, VD-NeRF incorporates a carefully designed training framework that enables robust camera pose estimation and self-supervised static-dynamic decomposition, allowing for more accurate reconstruction and perception of urban spatiotemporal scenes. The scene is divided into overlapping sub-scenes based on spatial dimensions, and training begins with a limited number of images, progressively incorporating more as training advances. Motion masks are utilized to minimize the interference of dynamic objects during camera pose estimation. Once accurate camera poses are established, the dynamic NeRF model reconstructs dynamic foregrounds and performs self-supervised decoupling of dynamic objects without relying on motion masks. By leveraging 3D scene flow, the model aggregates information across multiple time steps, enhancing the distinction between dynamic and static objects and reducing the likelihood of incorrect decoupling.

In summary, our core contributions are:
\begin{itemize}
    \item  We present a method that uses only image sequences from dynamic urban scenes to recover accurate camera trajectories, learn spatiotemporal scene representations, and achieve photorealistic dynamic novel view synthesis.
    \item  We propose a novel training framework that enhances the robustness and accuracy of camera pose estimation and urban scene reconstruction, enabling self-supervised static-dynamic decomposition without requiring precise annotations.
    \item Our method is extensively validated on mainstream urban scene datasets, demonstrating state-of-the-art performance in both camera pose estimation and dynamic novel view synthesis.
\end{itemize}

\section{Related Work}

\subsection{NeRF With Camera Pose Estimation.}
Eliminating camera pre-processing to achieve joint optimization of camera poses and NeRF is a hot research area. The first to utilize NeRF for optimizing camera poses is iNeRF \cite{yen2021inerf} which formulated camera pose estimation as an inverse problem. Leveraging a pre-trained NeRF model, it estimates poses by sampling key points. Since then, many NeRF-based pose-free methods have been proposed. NeRF\texttt{--} \cite{wang2021nerf} treats camera poses as learnable parameters, proposing a framework for joint optimization of camera poses and NeRF. GNeRF \cite{meng2021gnerf} introduces Generative Adversarial Networks (GAN) to optimize camera poses. BARF \cite{lin2021barf}extends the classical image alignment theory to NeRF, proposing a coarse-to-fine registration strategy for camera pose recovery. While recovering the camera poses,  SC-NeRF \cite{jeong2021self} also takes camera intrinsic and distortion as optimization parameters. L2G-NeRF \cite{chen2023local} introduces a local-to-global registration method to avoid falling into local minima. Nope-NeRF \cite{bian2023nope} utilizes monocular depth estimation to provide a prior to constrain the relative pose between consecutive images, recovering the large camera motion trajectory. LocalRF \cite{meuleman2023progressively} progressively jointly optimizes camera poses and NeRF, improving robustness in long camera trajectories and large-scale scenes. CF-NeRF \cite{yan2024cf} and CT-NeRF \cite{ran2024ct} use an incremental optimization approach to handle complex camera motion trajectories effectively. Although these methods have demonstrated remarkable performance in static scenes, they often face challenges in representing spatiotemporal urban scenes. This is because urban scenes typically feature long camera motion trajectories and dynamic objects, these methods struggle to accurately estimate camera poses and reconstruct dynamic objects.

Some NeRF-based methods estimate camera poses and reconstruct dynamic scenes without requiring input camera poses. RoDynRF \cite{liu2023robust} leverages 2D motion masks to jointly optimize camera poses and static and dynamic radiance fields. However, its accuracy is highly dependent on the quality of the motion masks and not be suitable for large-scale urban scenes. For omnidirectional dynamic images, OmniLocalRF \cite{choi2024omnilocalrf} proposed a bidirectional optimization strategy along with a motion mask prediction module to enable camera pose estimation and scene reconstruction in large-scale dynamic scenes. While it performs well in large-scale dynamic scenes, it struggles to effectively render or decouple dynamic objects. OmniLocalRF primarily focuses on removing and inpainting dynamic objects. In addition, its optimization strategy is specifically designed for omnidirectional images. In contrast, VDNeRF can simultaneously achieve accurate camera pose estimation and high-quality scene reconstruction in large-scale dynamic scenes such as urban scenes. In the absence of precise motion masks, it can also self-supervise static-dynamic decomposition and dynamic object rendering, thereby improving its performance for downstream applications.

\subsection{NeRF for Dynamic or Large-scale Scene.}
A fundamental limitation of NeRF is the assumption that the scene is strictly static, with constant volume density and emitted radiance (color). To address this, some works relax the static scene assumption of NeRF and extend it to dynamic scenes~\cite{gao2024cosurfgs, wu2025hv}. NeRF-W \cite{martin2021nerf} learns a latent embedding for each image to mitigate the effects of illumination variations and transient objects in the scene. D-NeRF \cite{pumarola2021d} introduces a temporal variable to learn the deformation field over different time instants. Nerfies \cite{park2021nerfies} proposes a continuous volumetric deformation field and a coarse-to-fine regularization method to ensure robust optimization. HyperNeRF \cite{park2021hypernerf} extends NeRF into a high-dimensional space, using slices of this space to reconstruct topologically variation scenes with discontinuous deformations. NSFF \cite{li2021neural} introduces forward and backward scene flow, along with multi-view constraints to achieve spatiotemporal view synthesis of in-the-wild scenes. DynamicNeRF \cite{gao2021dynamic} introduces multiple regularization losses to address the ambiguity in dynamic scene reconstruction. D$^{2}$NeRF \cite{wu2022d} achieves the decoupling of dynamic and static objects in scenes by employing two separate NeRF models to represent the scene. It also introduces a shadow field network specifically for handling situations like shadows, improving the rendering and decomposition quality of dynamic objects. DyBluRF \cite{sun2024dyblurf} achieves dynamic novel view synthesis from blurred dynamic videos by predicting the object's global Discrete Cosine Transform (DCT) trajectory. 

Beyond the challenge of handling dynamic scenes, another critical limitation of NeRF is its insufficient model capacity. Several methods have been proposed to extend NeRF to large-scale scenes. BungeeNeRF \cite{xiangli2022bungeenerf} reconstructs scenes at multi-scale by progressively growing the model and inclusive multi-level supervision strategies. Block-NeRF \cite{tancik2022block} and Mega-NeRF \cite{turki2022mega} employ multiple compact NeRF models for large-scale scene representation. S-NeRF \cite{xie2023s} and StreetSurf \cite{guo2023streetsurf} can render images from novel views in non-object-centric street-view scenes. In PreSight \cite{yuan2024presight}, a city-scale NeRF framework compatible with online perception models is constructed. 

Some works attempt to simultaneously address the limitations of NeRF in both static and small-scale settings to achieve urban scene reconstruction. SUDS \cite{turki2023suds} employs three scalable branches: static, dynamic, and far-field to represent city-scale dynamic scenes. Similar to SUDS, EmerNeRF \cite{yang2023emernerf} and RoDUS \cite{nguyen2024rodus} reconstruct urban scenes using multiple branches and decompose the scenes into dynamic and static components in a self-supervised manner. However, these methods heavily depend on accurate camera poses or even expensive LiDAR data, which are often difficult to obtain in real-world applications. Unlike methods such as Block-NeRF and SUDS, which focus on city-scale reconstruction by partitioning entire city scenes into blocks under known camera poses, VDNeRF is not specifically designed for city-scale scenes. However, it can jointly optimize camera poses and NeRF in smaller-scale urban scenes, such as a street. By creating sub-scenes based on the spatial extent of the estimated camera trajectory, VDNeRF can be scaled to arbitrarily long streets. Furthermore, using only a sequence of RGB images, VDNeRF can robustly reconstruct spatiotemporal scenes and decouple dynamic elements.

\begin{figure*}[ht]
    \centering
    \includegraphics[width=0.9\linewidth]{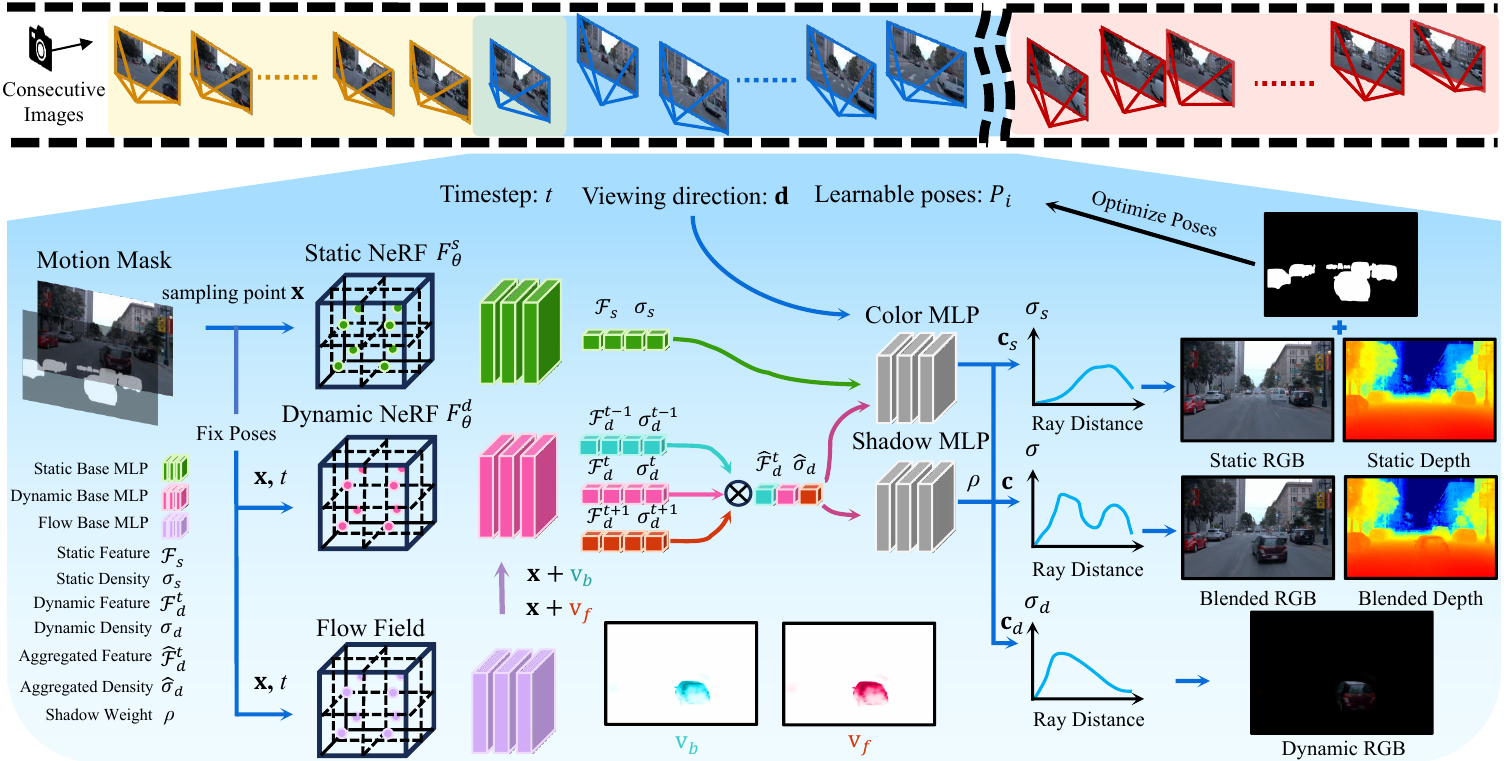}
    \caption{\textbf{VDNeRF Overview.} VDNeRF is composed of a static NeRF $F_{\Theta}^{s}$ and a dynamic NeRF $F_{\Theta}^{d}$ with a flow field. Each model is equipped with a hash grid and a base MLP. The static NeRF $F_{\Theta}^{s}$ takes the 3D location of the sampling points $\mathbf{x}=(x, y, z)$ as input, while dynamic NeRF $F_{\Theta}^{d}$ additionally incorporates a timestep $t$ as input. The static NeRF $F_{\Theta}^{s}$ query feature $\mathcal{F}_s$ and density $\sigma_s$ from the base MLP. For the dynamic NeRF $F_{\Theta}^{d}$, it leverages the 3D forward flow $\mathbf{v}_f$ and backward flow $\mathbf{v}_b$–with the 2D optical flow visualized in the figure–predicted by the flow field to aggregate dynamic feature $\hat{\mathcal{F}_d^t}$ and dynamic density $\hat{\sigma_d^t}$ across multiple timesteps and queries the shadow weight $\rho \in [0,1]$ from the shadow MLP. These features, along with the viewing direction $\mathbf{d}=(\theta, \phi)$, are passed to the color MLP to obtain the color $\mathbf{c}=(r, g, b)$. Volume rendering is applied to all sampling points along the camera ray to render static and dynamic pixel colors $\hat{C}_s(\mathbf{r})$ and $\hat{C}_d(\mathbf{r})$. Finally, the blended representation of spatiotemporal scenes is achieved using shadow weights. During training, VDNeRF partitions the scene into multiple overlapping sub-scenes. For each sub-scene, it starts with a small set of images, progressively optimizing the static NeRF $F_{\Theta}^{s}$ and camera poses $P_i$ with the help of motion masks. Once accurate camera poses are established, they are fixed, and the dynamic NeRF $F_{\Theta}^{d}$ is activated to reconstruct dynamic objects, achieving self-supervised static-dynamic decomposition.}
    \label{fig:method}
\end{figure*}
\section{Method}
VDNeRF jointly optimizes camera poses and NeRF from dynamic urban scenes in a self-supervised manner without camera poses and ground-truth depths. Given a sequence of images $\mathcal{I}=\left\{I_i \in \mathbb{R}^{H \times W \times 3} \mid i=0 \cdots N-1\right\}$ and focal length $f$, VDNeRF estimates camera poses $\mathcal{P}=\left\{P_i=\left[\mathbf{R}_i \mid \mathbf{t}_i\right] \in \mathrm{SE}(3) \mid i=0 \cdots N-1\right\}$, generates images $\mathcal{Y}_{r g b}=\left\{Y \in \mathbb{R}^{H \times W \times 3}\right\}$ from novel viewpoints, and decouple dynamic objects and static background.

In the following sections, we detail the workings of VDNeRF. Firstly, we briefly introduce NeRF in Section \ref{METHOD:A}. Next, we describe the architecture of our NeRF models in Section \ref{METHOD:B}, followed by an explanation of the regularization constraints for dynamic urban scenes in Section \ref{METHOD:C}. Then we propose our training framework in Section \ref{METHOD:D}, demonstrating how it enables robust camera pose estimation and dynamic object reconstruction in urban scenes. Finally, we discuss the loss functions that are minimized throughout the training process in Section \ref{METHOD:E}.
\subsection{NeRF Preliminary}
\label{METHOD:A}
Neural Radiance Fields (NeRF) \cite{mildenhall2021nerf} can implicitly model a static 3D scene and synthesize photorealistic novel views. It encodes the points in 3D scene as volume density $\sigma$ and emitted color $\mathbf{c}=(r, g, b)$ with an MLP network $F_{\Theta}:(\mathbf{x}, \mathbf{d}) \rightarrow(\mathbf{c}, \sigma)$, where $\mathbf{x}=(x, y, z)$ is a location of the 3D point and $\mathbf{d}=(\theta, \phi)$ is a 2D viewing direction, indicating the perspective from which the 3D point is observed. 

Specifically, given a posed image, along a camera ray $\mathbf{r}(t)=\mathbf{o}+t \mathbf{d}$ emitted from the camera optical center $\mathbf{o}$, NeRF samples $N$ 3D discrete points. Through the MLP network $F_{\Theta}$, volume density $\sigma$ and emitted color $\mathbf{c}$ of each sampling point can be obtained. Then using volume rendering \cite{kajiya1984ray}, NeRF can render one pixel color $\hat{C}(\mathbf{r})$:
\begin{equation}
\label{METHOD:eq1}
\hat{C}(\mathbf{r})=\sum_{i=1}^N T_i \alpha_i \mathbf{c}_i, T_i=\exp \left(-\sum\limits_{j=1}^{i-1} \sigma_j \delta_j\right),
\end{equation}
where $\alpha_i=1-\exp \left(-\sigma_i \delta_i\right)$ and $\delta_j=t_{i+1}-t_i$ is the distance between two sampling points. NeRF optimizes the MLP network $F_{\Theta}$ by minimizing the photometric error between synthesized pixels and ground truth observed images, enabling accurate scene representation.
\subsection{Static and Dynamic NeRF Models}
\label{METHOD:B}
Given the large number of objects and the extensive motion of dynamic elements in urban scenes, it is difficult to effectively represent such complex scenes using a single NeRF model. Therefore, following previous excellent works \cite{wu2022d,yang2023emernerf,nguyen2024rodus,turki2023suds}, VDNeRF uses two separate NeRF models (static NeRF $F_{\Theta}^{s}$ and dynamic NeRF $F_{\Theta}^{d}$) to represent the spatiotemporal scene. Each NeRF model employs learnable multi-resolution hash encoding as proposed by Instant-NGP \cite{muller2022instant} to enhance the speed of training and rendering. Static NeRF $F_{\Theta}^{s}$ reconstructs the background composed of non-moving objects. Dynamic NeRF $F_{\Theta}^{d}$ combines the 3D scene flow predicted by the flow field to reconstruct the foreground composed of moving objects. The two NeRF models query feature $\mathcal{F}$ and volume density $\sigma$ from the base MLPs. The feature $\mathcal{F}$ is then passed to the color MLP to obtain the corresponding color $\mathbf{c}$. Note that, to achieve the self-supervised decomposition of the scene, both NeRF models are trained for the entire scene. As a result, two NeRF models share a color MLP and collectively represent the spatiotemporal scene with the assistance of a shadow MLP. 

Static NeRF $F_{\Theta}^{s}$ is responsible for reconstructing the static background in the scene and optimizing camera poses $\mathcal{P}=\left\{P_i=\left[\mathbf{R}_i \mid \mathbf{t}_i\right] \in \mathrm{SE}(3) \mid i=0 \cdots N-1\right\}$:
\begin{equation}
\label{METHOD:eq2}
(\mathbf{c}_{s}, \sigma_{s})=F_{\Theta}^{s}(\mathbf{x}, \mathbf{d}) ,
\end{equation}
where $\mathbf{x}$ represents the location of a sampling point along a camera ray $\mathbf{r}(t)=\mathbf{o}+t \mathbf{d}$, and $\mathbf{d}$ denotes the direction of the camera ray $\mathbf{r}(t)$. The camera optical center $\mathbf{o}$ corresponds to the translation $\mathbf{t}_i \in \mathrm{T}(3)$, while the ray direction $\mathbf{d}$ is determined by the rotation matrix $\mathbf{R}_i \in \mathrm{SO}(3)$. Therefore, as static NeRF $F{\Theta}^{s}$ reconstructs the static background, it can simultaneously optimize the camera poses $\mathcal{P}$.

Dynamic NeRF $F_{\Theta}^{d}$ introduces an additional timestep $t$ as input, extending NeRF to the dynamic scene:
\begin{equation}
\label{METHOD:eq3}
(\mathbf{c}_{d}, \sigma_{d})=F_{\Theta}^{d}(\mathbf{x}, \mathbf{d},{t}) .
\end{equation}

However, relying solely on temporal encoding is insufficient to accurately reconstruct dynamic objects in urban scenes, such as fast-moving vehicles with long motion trajectories. This becomes especially apparent in the rendered novel views, where dynamic objects exhibit blurred shapes or even residual motion trails. To address this, dynamic NeRF $F_{\Theta}^{d}$ incorporates a flow field to predict the 3D scene flow. Given a timestep $t$ and a 3D location $\mathbf{x}=(x, y, z)$, it outputs a 3D flow vector $\mathbf{v} \in \mathbb{R}^3$ to predict the location of $\mathbf{x}$ in the adjacent timestep: $\mathbf{x}^{\prime}=\mathbf{x}+\mathbf{v}$. The flow field estimates the forward and backward 3D scene flows, $\mathbf{v}_f$ and $\mathbf{v}_b$, for each point. It ensures consistency between the forward flow at timestep $t$ and the backward flow at the next timestep $t+1$, then inputs these 3D scene flows into the dynamic NeRF $F_{\Theta}^{d}$. The 3D scene flows enable the dynamic NeRF $F_{\Theta}^{d}$ to better capture the motion of objects, facilitating more accurate and realistic rendering of dynamic objects from new timesteps or viewpoints. Additionally, following EmerNeRF, the dynamic NeRF $F_{\Theta}^{d}$ aggregate features $\hat{\mathcal{F}_d^t}$ and dynamic volume density $\hat{\sigma_d^t}$ from multiple timesteps:
\begin{equation}
\begin{split}
\label{METHOD:eq4}
(\mathbf{x}+\mathbf{v}_b, t-1) \rightarrow(\mathcal{F}_d^{t-1}, \sigma_d^{t-1}),
\\
(\mathbf{x}+\mathbf{v}_f, t+1) \rightarrow(\mathcal{F}_d^{t+1}, \sigma_d^{t+1}),
\\
\hat{\mathcal{F}_d^t}= \lambda_1 \cdot \mathcal{F}_d^{t-1} + \lambda_2 \cdot \mathcal{F}_d^{t} + \lambda_3 \cdot \mathcal{F}_d^{t+1},
\\
\hat{\sigma_{d}^t}=\lambda_1 \cdot  \sigma_d^{t-1} + \lambda_2  \cdot  \sigma_d^{t} + \lambda_3 \cdot \sigma_d^{t+1},
\end{split}
\end{equation}
where $\lambda_1=\lambda_3=0.25$ and $\lambda_2=0.5$.
It helps reduce the noise in temporal attributes and allows information to be shared across different timesteps, improving the rendering and decoupling quality of dynamic objects.

\begin{figure}[ht]
    \centering
    \includegraphics[width=0.9\linewidth]{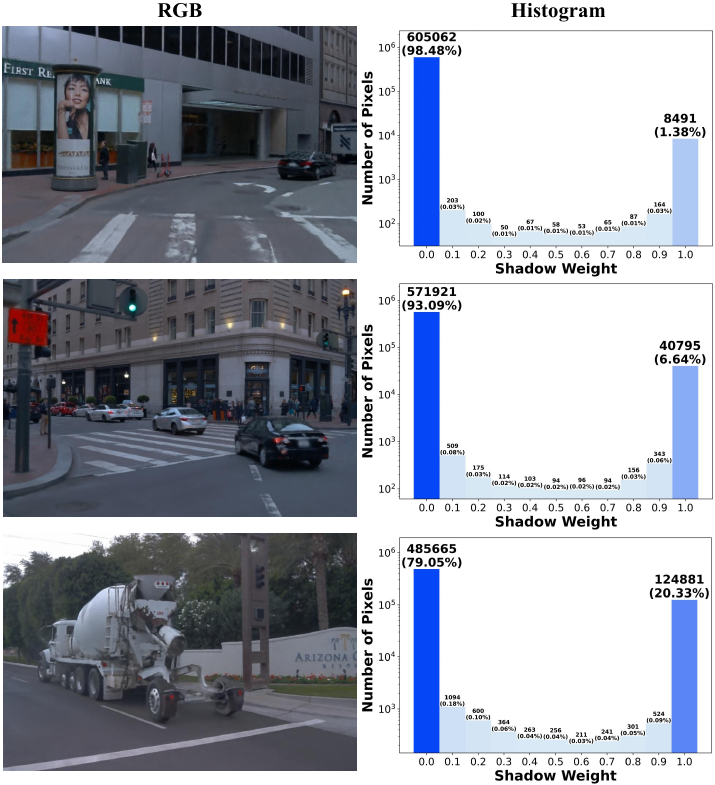}
    \caption{\textbf{Histograms of shadow weight distributions in different scenes.} The left column shows the synthesized RGB images, while the right column presents histograms of the shadow weight distributions corresponding to the pixels. The y-axis of each histogram is plotted on a logarithmic scale to better visualize the distribution. We also annotate each histogram with the exact counts and corresponding percentages. Please consider zooming in to view the detailed numbers. }
    \label{fig:histogram}
\end{figure}

Although static NeRF $F_{\Theta}^{s}$ and dynamic NeRF $F_{\Theta}^{d}$ are two separate NeRF models, they are fused when rendering the scene. As in the real-world environment, dynamic objects exist on a static background. However, due to the inherent complexity of dynamic scenes, such as motion blur, occlusion, or noise in dynamic object representations, we can not directly fuse the colors emitted from the dynamic NeRF $F_{\Theta}^{d}$ and the static NeRF $F_{\Theta}^{s}$. Inspired by D$^{2}$NeRF, we introduce a shadow MLP to control the proportion of the fusion of static NeRF $F_{\Theta}^{s}$ and dynamic NeRF $F_{\Theta}^{d}$. Unlike the original design, which primarily focuses on shadows for dynamic objects, we find that this approach is equally effective for the rendering and decomposition of the entire dynamic scene. This shadow MLP outputs a shadow weight $\rho \in [0,1]$ to downweight the static background. Finally, the emitted color $\hat{C}(\mathbf{r})$ of the sampling point can be formulated as:
\begin{equation}
\label{METHOD:eq5}
\mathbf{c}=(1-\rho) \cdot \frac{\sigma_s}{\sigma}\cdot \mathbf{c}_s + \frac{\sigma_d}{\sigma} \cdot \mathbf{c}_d,
\end{equation}
where volume density $\sigma$=$\sigma_s${+}$\sigma_d$. The shadow weight $\rho$ serves as a soft blending factor between the static NeRF $F_{\Theta}^{s}$ and dynamic NeRF $F_{\Theta}^{d}$, allowing VDNeRF to precisely represent static background and dynamic objects. Notably, the learned values of shadow weight $\rho$ reflect the dynamic content of a scene. When a scene contains more dynamic elements, more sampling points exhibit $\rho$ values close to 1; conversely, in scenes with fewer dynamic elements, most sampling points tend to have $\rho$ values close to 0. To more intuitively visualize the distribution of $\rho$ across different scenes, we render the shadow weights into images using volume rendering of NeRF, obtaining per-pixel $\rho$ values and plotting their histograms, as shown in Fig. \ref{fig:histogram}.
\subsection{Regularization Constraints for Urban scenes}
\label{METHOD:C}

In feed-forward and non-object-centric urban scenes, the observations of objects are transient and sparse. Under this setting, it is easy to trigger the inherent shape-radiance ambiguity \cite{zhang2020nerf++} of NeRF. This ambiguity resembles overfitting, preventing the scene reconstruction from generalizing to novel viewpoints. Unfortunately, jointly optimizing the camera poses and NeRF further exacerbates the shape-radiance ambiguity. Moreover, the reconstruction of dynamic objects is also affected by this ambiguity, as the significant variation in observed motion states across different timesteps allows for multiple possible fits. Depending exclusively on photometric error for supervision cannot mitigate this issue. To address this ambiguity, we introduce two regularization supervisions: monocular depth and 2D optical flow.

Using DPT \cite{ranftl2021vision}, we generate monocular depth maps from RGB images. These depth priors provide relative scale and geometric constraints that reduce convergence time and improve reconstruction quality from sparse viewpoints \cite{deng2022depth}. They also simplify the process of optimizing camera poses and effectively mitigate shape-radiance ambiguity \cite{bian2023nope}. Furthermore, they can help infer the direction of object motion, leading to a better reconstruction of dynamic objects.

The 2D optical flow generated by RAFT \cite{teed2020raft} also helps mitigate this ambiguity, enabling NeRF to plausibly represent challenging scenes \cite{li2021neural,gao2021dynamic}. By capturing pixel correspondences between adjacent images, optical flow imposes constraints on both the geometric structure of the scene and the motion relationships within it. This provides explicit priors for the motion trajectories of dynamic objects and the camera. Moreover, it indirectly supervises the 3D scene flow, enhancing the temporal consistency of dynamic objects.

Given that the monocular depth estimation and 2D optical flow priors generated by DPT and RAFT may contain errors, we adopt an annealing strategy, exponentially decaying their constraint weights during training.

For more details on the depth and optical flow supervision, please refer to Section \ref{METHOD:E}.

\begin{figure}[ht]
    \centering
    \includegraphics[width=0.9\linewidth]{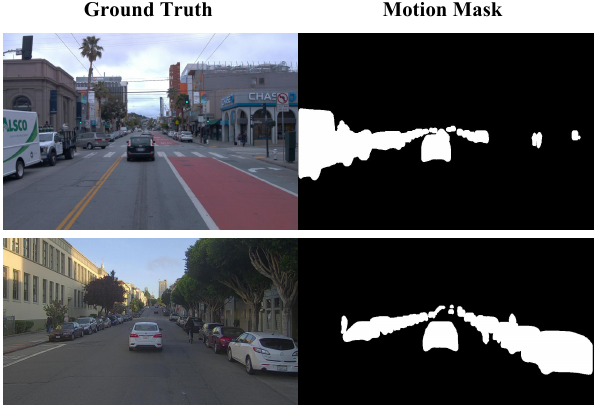}
    \caption{\textbf{Motion mask from RoDynRF.} The motion masks generated by RoDynRF include stationary vehicles and pedestrians on the roadside. We also do not use it to supervise static-dynamic decomposition but only use it to exclude dynamic elements to achieve more accurate and robust camera pose estimation. }
    \label{fig:mask}
\end{figure}
\subsection{Training Framework}
\label{METHOD:D}
Our training framework partitions the urban scene into multiple overlapping sub-scenes, each undergoing the following three training stages: (a) progressively optimizing static NeRF $F_{\Theta}^{s}$ and camera poses, (b) fixing the camera pose and activating dynamic NeRF $F_{\Theta}^{d}$ to model dynamic objects, and (c) dynamically creating a new sub-scenes. During camera pose optimization, we introduce motion masks from RoDynRF and compute the loss only in regions outside the motion masks, while keeping the dynamic NeRF $F_{\Theta}^{d}$ frozen. And the camera poses are fixed when $F_{\Theta}^{d}$ is activated. This separation is essential because distinguishing between pixel motion caused by dynamic objects and that caused by camera movement is challenging. Optimizing dynamic NeRF $F_{\Theta}^{d}$ and camera poses simultaneously in an urban scene easily leads the training in the wrong direction and falls into local minima, resulting in inaccurate camera pose estimation and failed dynamic object modeling. Consequently, this hinders the accurate representation and perception of the spatiotemporal scene. 

The three training stages are detailed as follows.

\textbf{Progressively Optimizing.} Even with the motion masks excluding the interference of dynamic objects, accurately estimating camera trajectory in narrow and long urban scenes remains a challenge. To ensure the accuracy and robustness of camera pose estimation, we adopt a progressive optimization approach inspired by LocalRF. Specifically, during the joint optimization of the static NeRF and camera poses, we do not input all images at once. Instead, it starts with only a small set of images and progressively incorporates additional images as the number of training iterations increases. When the number of images reaches a certain threshold or the camera trajectory exceeds a specified distance, no further images are added. At this point, rather than immediately activating the dynamic$F_{\Theta}^{d}$, we continue to jointly optimize static NeRF $F_{\Theta}^{s}$ and the camera poses while gradually reducing the learning rate for camera poses optimization to achieve more accurate camera pose estimation. The number of joint optimization iterations is determined by the total number of incorporated images.

\textbf{Activating Dynamic NeRF.} Once the accurate camera pose is established, we activate dynamic NeRF $F_{\Theta}^{d}$ and start to jointly optimize with the static NeRF $F_{\Theta}^{s}$ to achieve a spatiotemporal representation of the urban scene. Notably, unlike RoDynRF, we do not treat the motion mask as the definitive dynamic region, as it often misclassifies stationary pedestrians and parked vehicles as dynamic objects in urban scenes (see in Fig. \ref{fig:mask}). Instead, the static-dynamic decomposition of the scene relies only on the self-supervision of VDNeRF. This is made possible by the ability of our dynamic NeRF $F_{\Theta}^{d}$ to model object motion accurately and consistently. Additionally, by aggregating dynamic information across multiple timesteps, we effectively reduce noise, further highlighting the distinction between dynamic and static objects. To reinforce this decomposition process, we also impose constraints on the dynamic volume density $\hat{\sigma_d^t}$, encouraging it to be generated only when necessary. It's just that in the motion mask, we allow it to generate dynamic volume density $\sigma_{d}$ more easily and assume that the dynamic colors $\hat{C}^{d}(\mathbf{r})$ is more reliable when rendering the scene. We chose motion masks from RoDynRF simply because they can be conveniently and quickly generated in a self-supervised manner. However, we believe that if there is a more accurate motion mask, VDNeRf will perform better in static-dynamic decomposition and camera pose estimation. Because the motion mask contains parked vehicles, etc, which have obvious geometric and texture features, it can improve the accuracy of camera pose estimation and the quality of dynamic scene representation to some extent.

\textbf{Creating a New Sub-scenes.} When the training of the current sub-scene is complete, we create a new sub-scene and a new NeRF model. However, instead of discarding all images from the current NeRF model, we pass the last few images along with their camera poses to the next sub-scene. The scene depicted by these images will be jointly reconstructed by both NeRF models. Dividing the elongated urban scene into overlapping sub-scenes along spatial dimensions ensures that the optimization starts in the correct direction, maintaining the consistency and continuity of both the camera trajectory and object motion. Additionally, the strategy of using multiple NeRF models to represent large-scale urban scenes helps overcome the capacity limitations of a single NeRF model, thereby improving reconstruction quality.
\begin{figure*}[ht]
    \centering
    \includegraphics[width=0.95\linewidth]{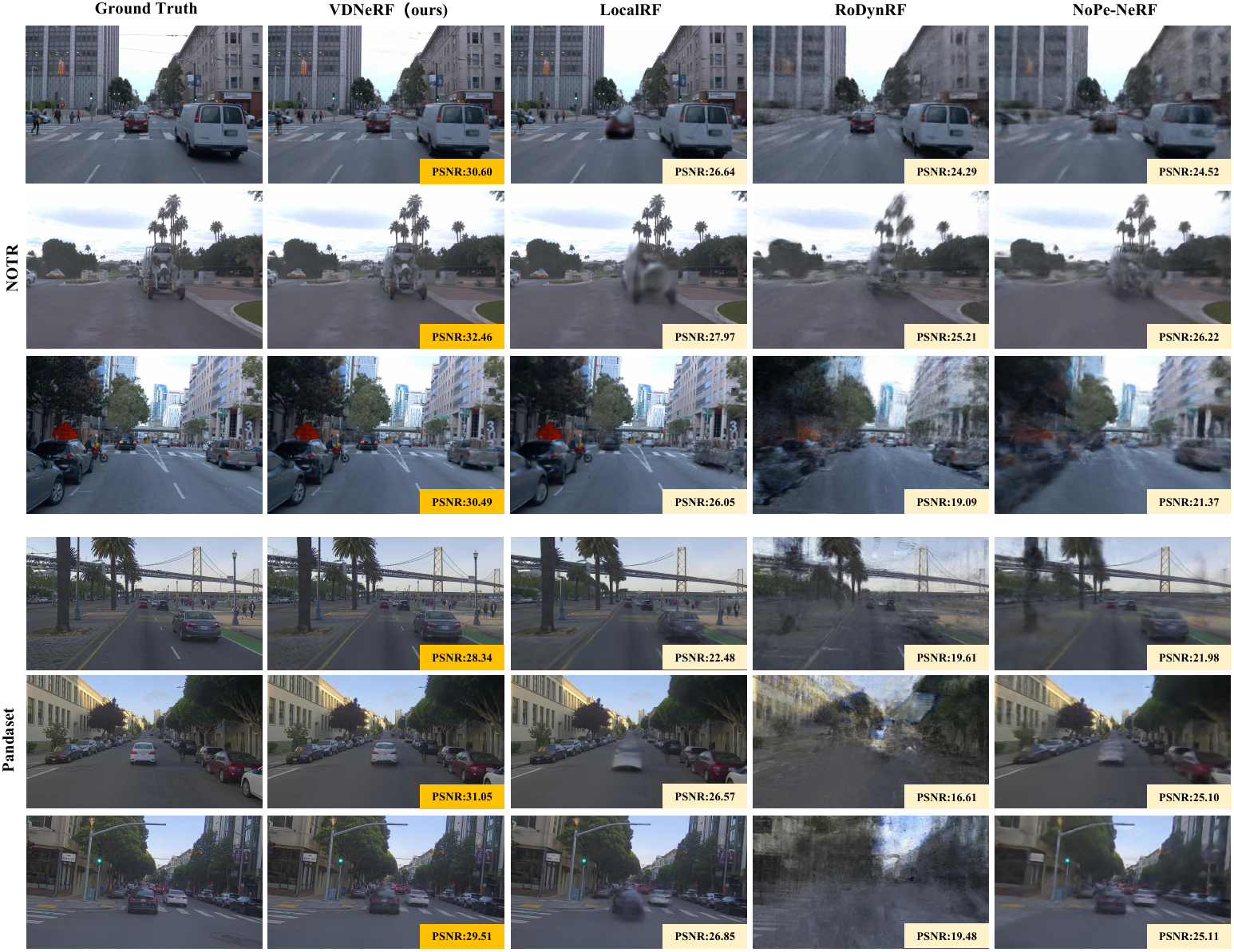}
    \caption{\textbf{Qualitative results of dynamic novel view synthesis on the NOTR and Pandaset datasets.} We visualize the images rendered from novel viewpoints. Compared to other methods, VDNeRF accurately reconstructs dynamic objects and synthesizes more detailed and photorealistic images.}
    \label{fig:nvs_notr}
\end{figure*}
\subsection{Loss Functions}
\label{METHOD:E}
\noindent\textbf{Total loss.} The total loss for VDNeRF is:
\begin{equation}
\begin{split}
\label{METHOD:eq13}
\mathcal{L}_{\text{total}}=\underbrace{\mathcal{L}_{\text{color}}+\mathcal{L}_{\text{depth}}+\mathcal{L}_{\text{flow}}}_{\text {also for camera pose }}\\+\underbrace{\mathcal{L}_{\text{cycle}}+\mathcal{L}_{\text{dynamic}}+\mathcal{L}_{\text{shadow}}}_{\text {when dynamic NeRF is activated}}.
\end{split}
\end{equation}

When we jointly optimize the camera poses and the static NeRF $F_{\Theta}^{s}$, we only use $\mathcal{L}_{\text{color}}$, $\mathcal{L}_\text{{depth}}$ and $\mathcal{L}_{\text{flow}}$. When we activate the dynamic NeRF $F_{\Theta}^{d}$, we enable $\mathcal{L}_{\text{cycle}}$, $\mathcal{L}_{\text{dynamic}}$ and $\mathcal{L}_{\text{shadow}}$.Below, we provide a detailed explanation of each loss function.

\noindent\textbf{Color loss.} We minimize the photometric error between rendered pixel color $\hat{C}(\mathbf{r})$ and the ground truth $C(\mathbf{r})$:
\begin{equation}
\label{METHOD:eq6}
\mathcal{L}_{\text{color}}=\left\|\hat{C}(\mathbf{r})-C(\mathbf{r})\right\|_2^2.
\end{equation}
\textbf{Depth loss.} We utilize the monocular depth estimation $D(\mathbf{r})$ from DPT \cite{ranftl2021vision} as a prior. Similar to \eqref{METHOD:eq1}, NeRF can render the depth of pixel:
\begin{equation}
\label{METHOD:eq7}
\hat{D}(\mathbf{r})=\sum_{i=1}^N T_i \alpha_i d_i, T_i=\exp \left(-\sum\limits_{j=1}^{i-1} \sigma_j \delta_j\right),
\end{equation}
where $d_i$ is the distance from the sampling point to the camera origin, $N$ is the number of sampling points on the camera ray corresponding to the pixel, $\delta_j=t_{i+1}-t_i$ represents the distance between two sampling points, $\alpha_i=1-\exp \left(-\sigma_i \delta_i\right)$. Since monocular depth estimation is not scaled and shift-invariant, we normalize $\hat{D}(\mathbf{r})$ and $D(\mathbf{r})$ to obtain normalized depth $\hat{D}^*(\mathbf{r})$ and $D^*(\mathbf{r})$ \cite{ranftl2020towards}. So the depth supervision is:
\begin{equation}
\label{METHOD:eq8}
\mathcal{L}_{\text{depth}}=\left\|\hat{D}^*(\mathbf{r})-D^*(\mathbf{r})\right\|_2^2.
\end{equation}
\textbf{Flow loss.} We use RAFT \cite{teed2020raft} to introduce 2D optical flow supervision $\mathbf{f}_{i \rightarrow i-1}$ and $\mathbf{f}_{i \rightarrow i+1}$. Using the depth and adjacent camera pose, we project a pixel into a 3D point and then reproject it onto adjacent images to compute the expected backward optical flow $\hat{\mathbf{f}}_{i \rightarrow i-1}$ and forward optical flow $\hat{\mathbf{f}}_{i \rightarrow i+1}$. We minimize the error between them and the optical flows estimated by RAFT:
\begin{equation}
\label{METHOD:eq9}
\mathcal{L}_{\text{flow}}=\left\|\hat{\mathbf{f}}_{i \rightarrow i-1}-\mathbf{f}_{i \rightarrow i-1}\right\|_1+\left\|\hat{\mathbf{f}}_{i \rightarrow i+1}-\mathbf{f}_{i \rightarrow i+1}\right\|_1.
\end{equation}
\textbf{Cycle loss.} As in previous work \cite{li2021neural,turki2023suds,yang2023emernerf}, for the 3D scene flow predicted by flow field, we also introduce a 3D scene flow cycle consistency term to encourage cycle consistency between forward and backward 3D scene flow:
\begin{equation}
\begin{split}
\label{METHOD:eq10}
\mathcal{L}_{\text{cycle}}=(\mathbf{v}_f(\mathbf{x},t_i)+\mathbf{v}_b(\mathbf{x}+\mathbf{v}_f(\mathbf{x},t_i),t_{i+1}))^2
\\+(\mathbf{v}_b(\mathbf{x},t_i)+\mathbf{v}_f(\mathbf{x}+\mathbf{v}_b(\mathbf{x},t_i),t_{i-1}))^2.
\end{split}
\end{equation}
\textbf{Dynamic loss.} In real-world urban scenes, dynamic objects are a minority. Therefore, to better decompose dynamic objects, we penalize the dynamic volume density $\sigma_{d}$ to encourage them to be generated only when necessary:
\begin{equation}
\label{METHOD:eq11}
\mathcal{L}_{\text{dynamic}}=\frac{1}{N}\sum_{i=1}^N \sigma_d(\mathbf{x}, t),
\end{equation}
where $N$ is the number of sampling points on the camera ray corresponding to the pixel.

\noindent\textbf{Shadow loss.} Follow D$^{2}$NeRF \cite{wu2022d}, to avoid over-interpreting dark regions of the scene and ensure that the scene is composed of a static background as much as possible, we penalize the square of the shadows ratio $\rho$ along a camera ray:
\begin{equation}
\label{METHOD:eq12}
\mathcal{L}_{\text{shadow}}=\sum_{i=1}^N T_i \alpha_i \rho_i^2.
\end{equation}

\renewcommand{\arraystretch}{1.2} 
\begin{table*}[ht]
\centering
\caption{\textbf{Quantitative results of dynamic novel view synthesis on the NOTR dataset.} We report PSNR, SSIM, and LPIPS of novel view synthesis. $\uparrow$: higher is better, $\downarrow$: lower is better. The \colorbox{red}{red}, \colorbox{orange}{orange}, and \colorbox{yellow}{yellow} highlights respectively denote the best, the second best, and the third best results. VDNeRF significantly leads other methods in all metrics.}
\setlength{\tabcolsep}{0.61mm}{
\footnotesize
\begin{tabularx}{\textwidth}{cc c ccc c ccc c ccc c ccc c ccc}
\toprule 
 & \multirow{3}{*}{\normalsize Scenes} &\vline&\multicolumn{15}{c}{\normalsize  \ding{51}  \ pose-free}&\vline&\multicolumn{3}{c}{\normalsize \ding{55} \ pose-free} 
 \\\cline{4-22} &&\vline&\multicolumn{3}{c}{Ours}  && \multicolumn{3}{c}{LocalRF} &&   \multicolumn{3}{c}{NoPe-NeRF} &&   \multicolumn{3}{c}{RoDynRF}  &\vline&  \multicolumn{3}{c}{EmerNeRF} 
 \\ \cline{4-6} \cline{8-10} \cline{12-14} \cline{16-18}  \cline{20-22} &&\vline& PSNR $\uparrow$ & SSIM $\uparrow$ & LPIPS $\downarrow$ && PSNR $\uparrow$ & SSIM $\uparrow$ & LPIPS $\downarrow$ &&  PSNR $\uparrow$ & SSIM $\uparrow$ & LPIPS $\downarrow$ & &  PSNR $\uparrow$ & SSIM $\uparrow$ & LPIPS $\downarrow$ &\vline& PSNR $\uparrow$ & SSIM $\uparrow$ & LPIPS $\downarrow$ 
 \\
\hline
\multicolumn{1}{c}{\multirow{4}{*}{\rotatebox[origin=c]{90}{NOTR}}} & 016 && \best 33.133 & \best0.920 & \best0.186 & & \tbest28.036 &\tbest 0.860  & \tbest0.282  && 25.001   & 0.701 & 0.482   && 23.810 & 0.678 & 0.475&\vline& \sbest30.928 & \sbest0.876 & \sbest0.270 \\
\multicolumn{1}{c}{}& 021 && \best30.902 & \best0.879 & \best0.292 & &\tbest 26.391 & \tbest0.843  & \sbest0.333  && 25.666   & 0.793 & 0.443   && 22.706 & 0.721 & 0.447&\vline& \sbest28.387 & \sbest0.850 & \tbest0.335 \\
\multicolumn{1}{c}{}& 022 && \best32.250 & \best0.913 & \best0.219 & &\tbest 28.145 &\tbest 0.862  & \tbest0.298  && 24.362   & 0.711 & 0.485    && 20.406 & 0.624 & 0.546&\vline& \sbest29.814 & \sbest0.880 & \sbest0.269 \\
\multicolumn{1}{c}{}& 025 && \best30.998 & \best0.887 & \best0.227 & &\tbest 27.419 & \tbest0.836  & \tbest0.300  && 23.832   & 0.651 & 0.520     && 24.186 & 0.677 & 0.454&\vline& \sbest29.685 & \sbest0.850 & \sbest0.280 \\
\hline
\multicolumn{1}{c}{\multirow{3}{*}{\rotatebox[origin=c]{90}{\scriptsize{Pandaset}}}} & 011 && \best27.202 & \best0.814 & \best0.260 & & \tbest22.729 & \sbest0.731  & \sbest0.361 && 21.768   & 0.632 & 0.503   && 19.980 & 0.558 & 0.539&\vline& \sbest23.072 & \tbest0.681 &\tbest 0.396 \\
\multicolumn{1}{c}{}& 027 && \best30.527 & \best0.878 & \best0.239 & &\tbest 26.185 & \sbest0.793  & \sbest0.363  && 24.592   & 0.685 & 0.498    && 16.215 & 0.522 & 0.611&\vline& \sbest26.231 & \tbest0.746 & \tbest0.413 \\
\multicolumn{1}{c}{}& 047 && \best31.090 & \best0.883 & \best0.199 & & \sbest26.906 & \sbest0.787  & \sbest0.333  && 25.060   & 0.664 & 0.503    && 16.142 & 0.488 &0.615&\vline& \tbest26.859 & \tbest0.731 & \tbest0.395 \\ 
\bottomrule
\end{tabularx}
}
\label{table:nvs}
\end{table*}
\begin{figure}[ht]
    \centering
    \includegraphics[width=0.95\linewidth]{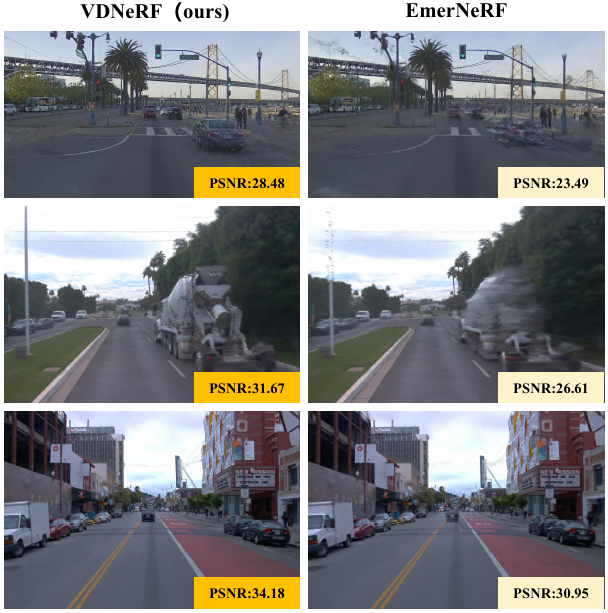}
    \caption{\textbf{Novel view synthesis comparison of VDNeRF and EmerNeRF.} EmerNeRF takes ground-truth camera poses as input but does not use ground-truth depths. The qualitative results of dynamic novel view synthesis demonstrate that VDNeRF achieves superior rendering quality with finer details and more accurate dynamic object modeling.}
    \label{fig:emer}
\end{figure}
\section{Experiments}
We first introduce our experimental setup in Section \ref{EXPERIMENTS:A}. Next, in Section \ref{EXPERIMENTS:B} and Section \ref{EXPERIMENTS:C}, we evaluate VDNeRF from two perspectives: novel view synthesis and camera pose estimation respectively, and compare it against mainstream NeRF-based methods. Then, we conduct further ablation studies in Section \ref{EXPERIMENTS:D}. Finally, we summarize the limitations of VDNeRF in Section \ref{EXPERIMENTS:E}.
\subsection{Experimental Setup}
\label{EXPERIMENTS:A}
\textbf{Dataset.} We selected two mainstream urban driving datasets, NeRF On-The-Road(NOTR) \cite{yang2023emernerf} and Pandaset \cite{xiao2021pandaset} to evaluate our method. The scenes we selected all contain certain dynamic objects. \textbf{NOTR:} this dataset is derived from the Waymo Open Dataset \cite{sun2020scalability}, which can reflect the challenges of training NeRF in real-world urban driving scenes. We use four dynamic scenes, each with 199 consecutive images captured by the front camera. All images are down-sampled to 960×640. Excluding the first image, one out of every ten images is selected for novel view synthesis. $\mathbf{Pandaset:}$ is also a challenging urban driving dataset. We use three dynamic scenes, each containing 80 consecutive images captured by the front-facing camera. All images are down-sampled to 960×540. The image selection strategy for novel view synthesis follows that of NOTR, except for the first image, every tenth image is selected.

\textbf{Comparisons with SOTA Methods.} We primarily compare VDNeRF against RoDynRF \cite{liu2023robust}, LocalRF \cite{meuleman2023progressively}, NoPe-NeRF \cite{bian2023nope}, as these methods are state-of-the-art NeRF-based pose-free methods.  Our evaluation focuses on two key aspects: novel view synthesis and camera trajectory recovery. In addition, RoDynRF also extends NeRF to dynamic scenes. Therefore, we further compare it in terms of dynamic object rendering quality and static-dynamic decomposition effectiveness. All baseline methods are configured and executed according to their official open-source implementations. To further evaluate the performance of VDNeRF in dynamic urban scenes, we compare it against the state-of-the-art NeRF-based method EmerNeRF \cite{yang2023emernerf}. It is important to note that EmerNeRF inputs the ground truth camera poses but does not input LiDAR point clouds. In the NOTR dataset, EmerNeRF inputs the motion masks and sky masks that are provided by the dataset, whereas such masks are not available in the Pandaset dataset.
\begin{figure*}[ht]
    \centering
    \includegraphics[width=0.9\linewidth]{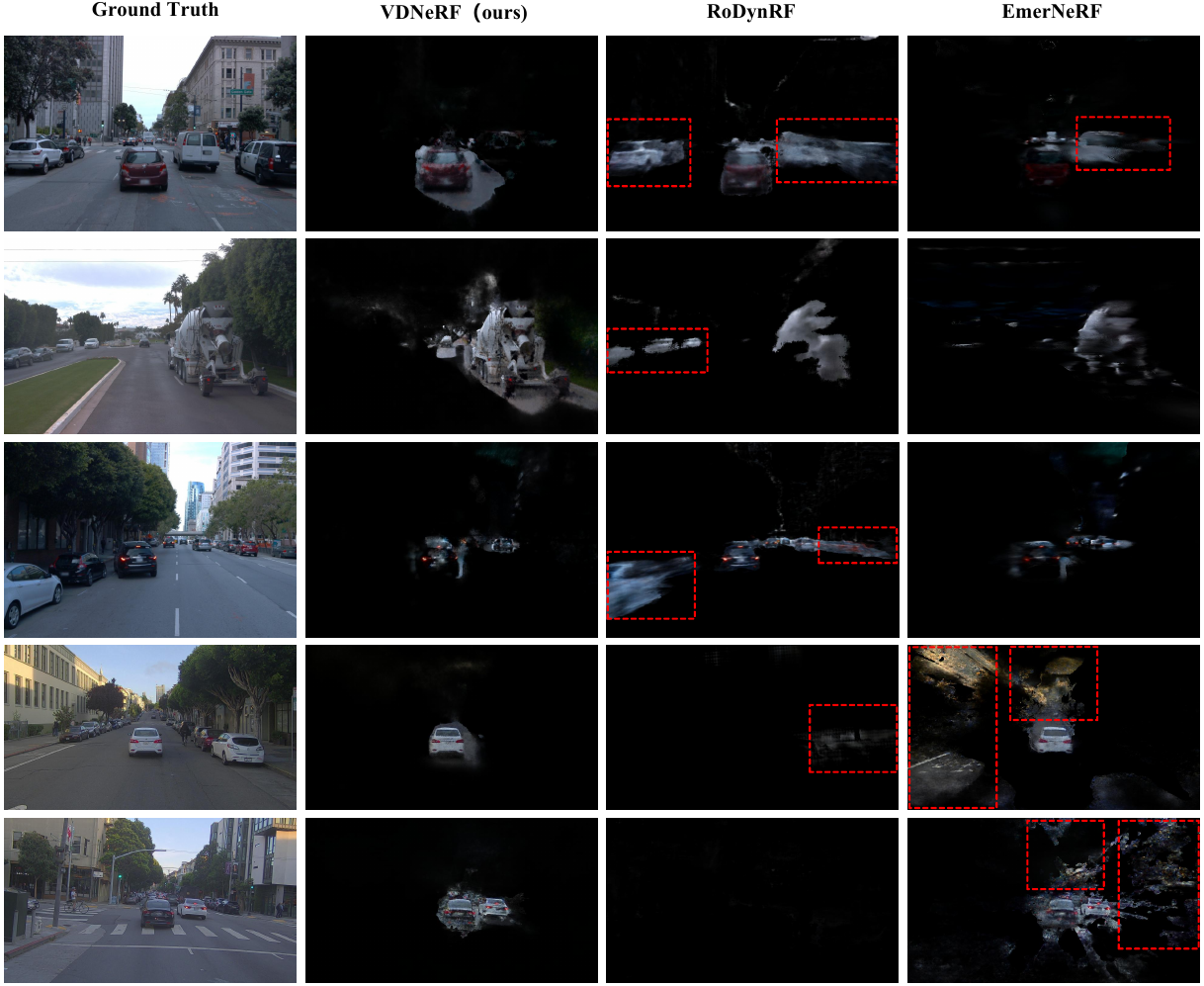}
    \caption{\textbf{Qualitative results of dynamic decoupling.} We visualize the dynamic RGB rendered by RoDynRF, EmerNeRF, and VDNeRF. VDNeRF outperforms other methods in terms of both the accuracy of static-dynamic decomposition and the rendering quality of dynamic objects. In contrast, RoDynRF and EmerNeRF occasionally misclassify static backgrounds as dynamic objects (see the red box in the image) or struggle to accurately render and separate dynamic objects.}
    \label{fig:motion}
\end{figure*}

\textbf{Metrics.} For the rendered images from novel views, we employ three standard error metrics: Peak Signal-to-Noise Ratio (PSNR), Structural Similarity Index Measure (SSIM) \cite{wang2004image}, and Learned Perceptual Image Patch Similarity (LPIPS) \cite{zhang2018unreasonable}. PSNR intuitively reflects the errors of each pixel. SSIM takes into account the adjacent pixels, focusing on the structure of the image. LPIPS uses a pre-trained neural network model (VGG architecture) to compare the deep neural information of the image, which is more consistent with human perception. For the recovered camera trajectory, we first use Umeyama \cite{umeyama1991least} to align it with the ground truth. We then evaluate it using two standard metrics: Absolute Trajectory Error (ATE) and Relative Pose Error (RPE). ATE measures the camera position errors of the entire camera trajectory and the ground truth. RPE measures the errors between pairs of camera poses, including relative rotation error (RPE$_r$), and relative translation error (RPE$_t$). 

\textbf{Implementation Details.} We build VDNeRF based on Instant-NGP \cite{muller2022instant} with the tiny-cuda-nn \cite{muller2021tiny} and use NerfAcc \cite{li2022nerfacc} to accelerate. All parameters are optimized by Adam optimizer \cite{kingma2014adam}. The train starts with 5 images. In each iteration, we render a total of 4096 pixels from different images. During the progressive optimization, we incorporate the new image every 600 iterations. To parameterize the unbounded scene, we adopt the sampling strategy of Mip-NeRF 360 \cite{barron2022mip}. When the head-to-tail distance of the estimated camera trajectory exceeds 4 or the total number of images contained surpasses 70, we stop adding new images and begin refining the current sub-scene. In urban driving scenes, these thresholds are chosen to create sub-scenes of appropriate spatial scale, enabling high-quality spatiotemporal reconstruction. The number of refinement iterations $N$ is determined by the total number of images, following a schedule of 840 iterations per incorporated image. In the first seventh of iterations $N$, we jointly optimize the static NeRF $F_{\Theta}^{s}$ and the camera poses. During this phase, the learning rate of camera poses and the regularization weights of flow and depth decrease exponentially to their original 0.1 factor. Afterward, we activate dynamic NeRF $F_{\Theta}^{d}$, at which point the learning rate of NeRF also decreases exponentially to its original 0.1 factor. Once the current sub-scene is fully trained, we incorporate the last twenty images into a created new sub-scene and repeat the above process until all images are trained. Training one scene (199 images) of the NOTR dataset takes approximately 12 hours on an NVIDIA RTX3090 GPU. The training proceeds at about 7 iterations per second, with a maximum GPU memory usage of around 12 GB. Synthesizing a blended RGB image (960×640) takes approximately 2.7 seconds.
\begin{figure*}[ht]
    \centering
    \includegraphics[width=0.95\linewidth]{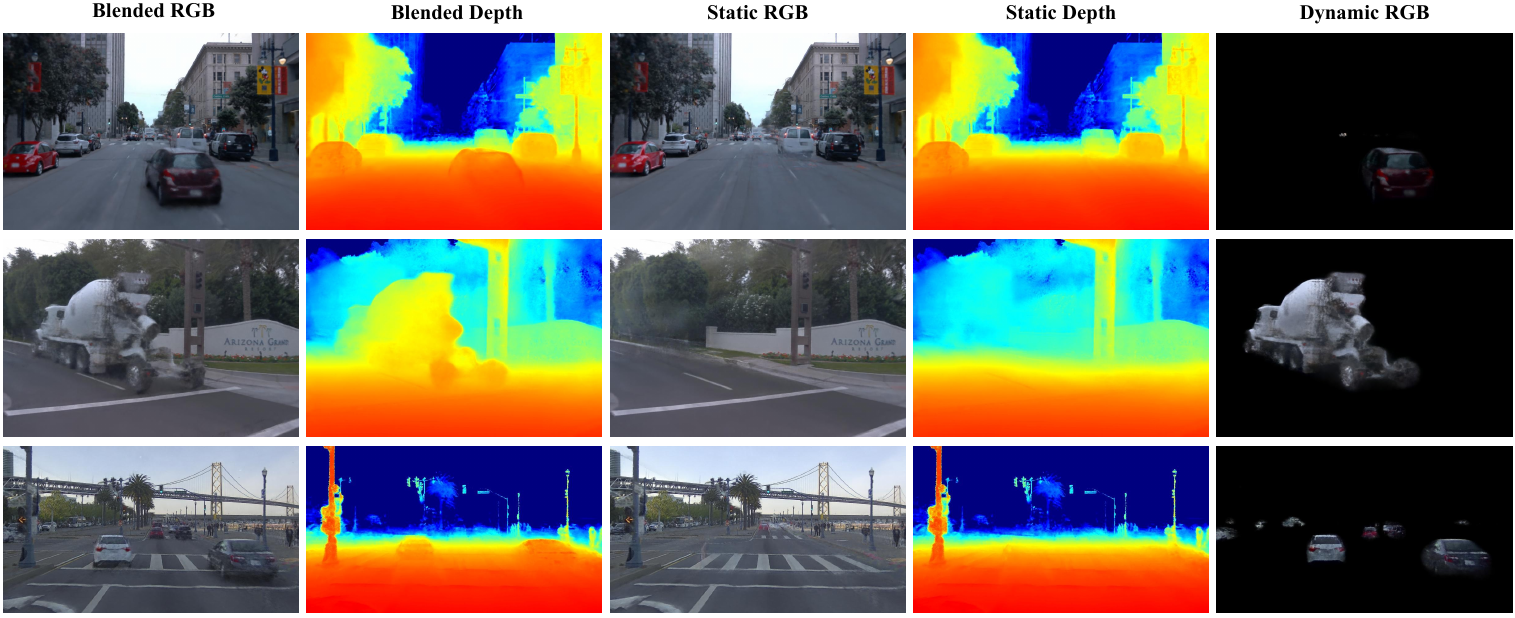}
    \caption{\textbf{Our results of static-dynamic decomposition.} We visualize the blended RGB and depth, as well as the static RGB, static depth, and dynamic RGB for a more comprehensive presentation.}
    \label{fig:decomposition}
\end{figure*}
\begin{figure*}[ht]
    \centering
    \includegraphics[width=0.95\linewidth]{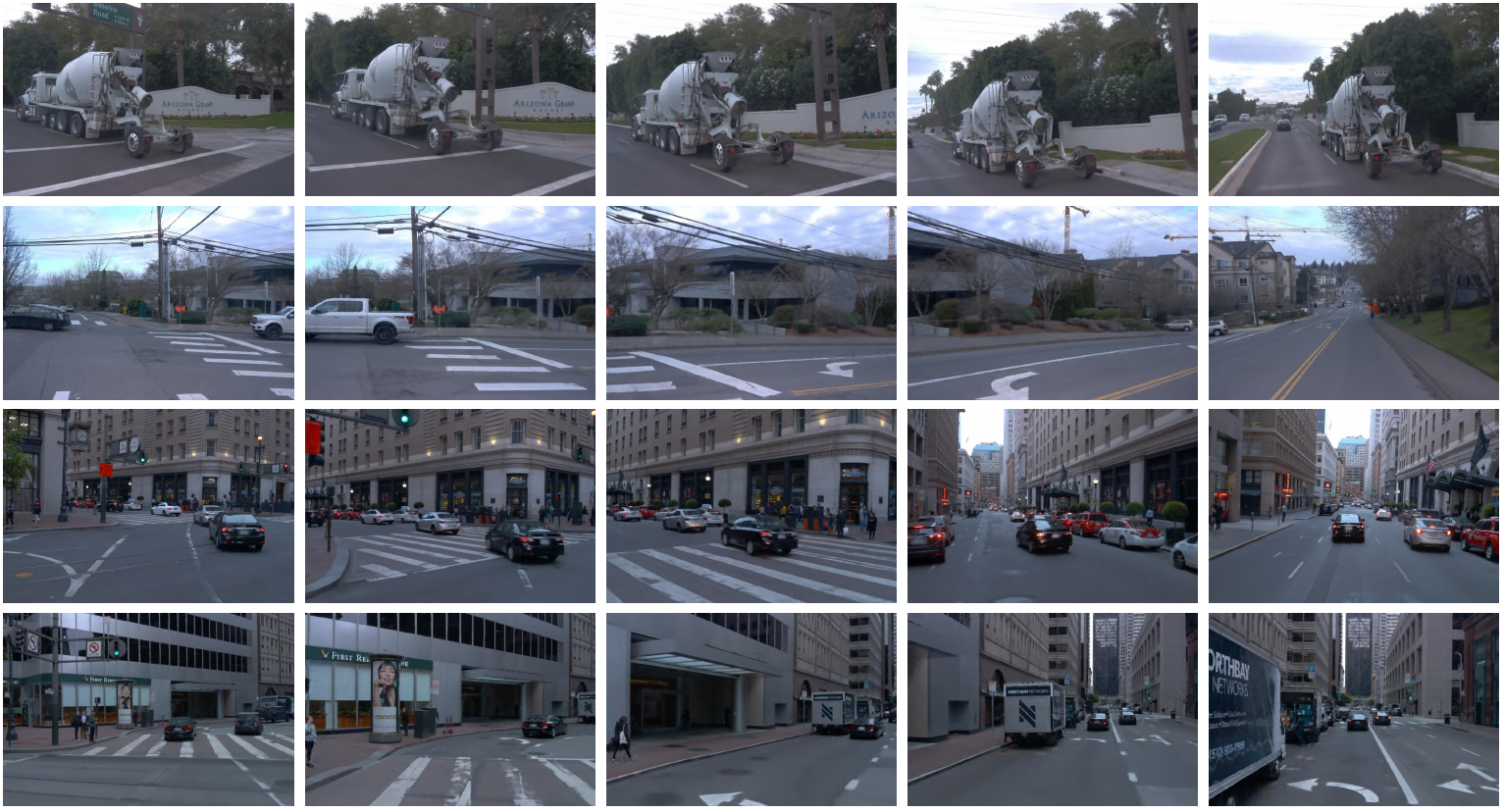}
    \caption{\textbf{Visualization of turning scenarios.} We present synthesized images during turning maneuvers to demonstrate that VDNeRF is capable of accurately reconstructing turning scenarios.}
    \label{fig:turn}
\end{figure*}
\subsection{Comparison of Novel View Synthesis}
\label{EXPERIMENTS:B}
Following the post-process strategy of NeRF-based pose-free methods like Nope-NeRF and LocalRF, we obtain the camera poses for test viewpoints by freezing the NeRF model and optimizing only the camera poses. Using these learned camera poses, we synthesize novel view images for testing. We present the quantitative results of novel view synthesis in TABLE \ref{table:nvs}. Across two different datasets, VDNeRF obtains consistent and significant gains across all metrics. Compared to the second-best method, VDNeRF achieves average improvements of 2.117 in PSNR, 0.036 in SSIM, and a drop of 0.057 in LPIPS on the NOTR dataset. On the Pandaset dataset, VDNeRF achieves average improvements of 4.203 in PSNR, 0.088 in SSIM, and a drop of 0.120 in LPIPS.

For an intuitive comparison, Fig. \ref{fig:nvs_notr} shows the qualitative results of novel view synthesis for the evaluated methods. The novel view images synthesized by VDNeRF have clearer and more realistic geometry and texture. Notably, in the absence of ground-truth depths, VDNeRF performs even better than the state-of-the-art methods for dynamic driving scenes, EmerNeRF, which relies on ground-truth camera poses. Qualitative comparisons are shown in Fig. \ref{fig:emer}.

LocalRF and NoPe-NeRF are designed for static scenes. While they can sometimes fit a roughly correct scene by leveraging their scene representation capabilities, they fail to accurately reconstruct the motion and geometry of dynamic objects. 

Both RoDynRF and EmerNeRF are capable of modeling and rendering dynamic objects to achieve static-dynamic decomposition. Therefore, VDNeRF also compares with RoDynRF and EmerNeRF for the dynamic decoupling performance in the synthesized novel view images. Qualitative results are displayed in Fig. \ref{fig:motion}. Unlike VDNeRF and EmerNeRF, which are self-supervised, RoDynRF relies on generated motion masks as supervision. However, in urban scenes, these motion masks often misclassify parked vehicles as dynamic objects, causing RoDynRF to incorrectly model stationary objects as dynamic ones, leading to inaccurate scene representation. Additionally, the model capacity of RoDynRF is limited, leading to low-resolution reconstructions when handling large-scale urban scenes. Without input LiDAR point clouds, although EmerNeRF can achieve static-dynamic decomposition, its decomposition accuracy and the rendering quality of dynamic objects are not as good as VDNeRF.

We present the RGB images obtained from our static-dynamic decomposition along with the corresponding depth maps in Fig. \ref{fig:decomposition}. Moreover, VDNeRF is capable of generating high-quality spatiotemporal scene representations even under rotational camera motion, such as intersection turns. The visualization of turning scenarios is presented in Fig. \ref{fig:turn}.

\begin{figure*}[ht]
    \centering
    \includegraphics[width=0.95\linewidth]{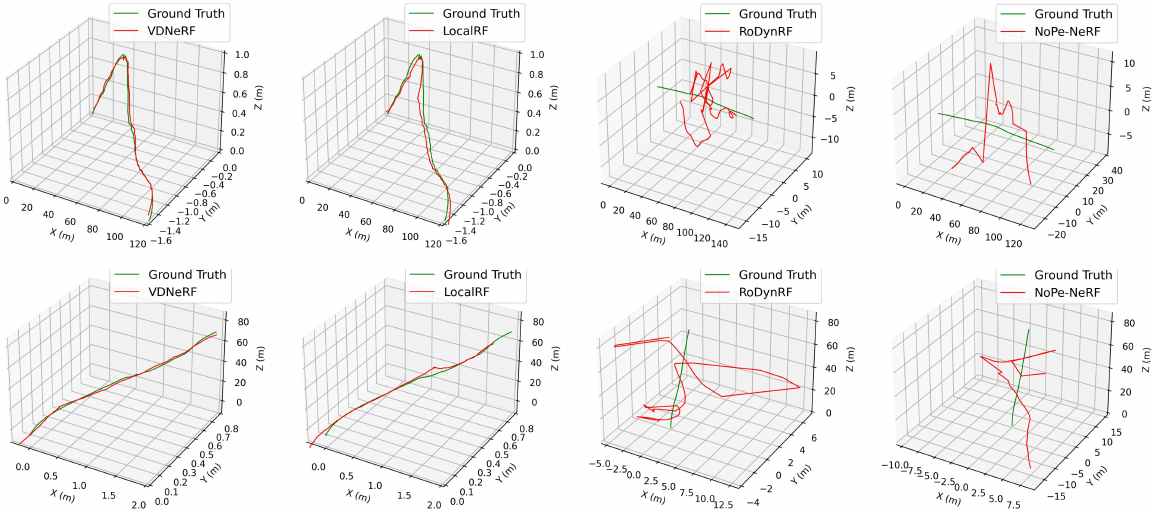}
    \caption{\textbf{Qualitative results of camera pose estimation.} For ease of comparison, VDNeRF and LocalRF are visualized within the same coordinate system range. Compared to other NeRF-based pose-free methods, VDNeRF estimates camera trajectories that align more closely with the ground truth.}
    \label{fig:pose}
\end{figure*}
\begin{table*}[ht]
\centering
\caption{\textbf{Quantitative comparison of camera pose estimation on the NOTR and Pandaset datasets.} We report $\text{RPE}_t$, $\text{RPE}_r$ and ATE of the recovered camera trajectory. The unit of $\text{RPE}_r$ is in degrees, while $\text{RPE}_t$ and ATE is in the ground truth scale. $\uparrow$: higher is better, $\downarrow$: lower is better. The \colorbox{red}{red}, \colorbox{orange}{orange}, and \colorbox{yellow}{yellow} highlights respectively denote the best, the second best, and the third best results. The camera trajectories recovered by VDNeRF have better accuracy and global consistency.}
\small
\setlength{\tabcolsep}{1.45mm}{
\begin{tabularx}{0.95\textwidth}{cccccccccccccccccc} 
\toprule 
\multirow{2}{*}{} & \multirow{2}{*}{scenes} &\vline& \multicolumn{3}{c}{Ours}                                                 &  & \multicolumn{3}{c}{LocalRF}                   &  & \multicolumn{3}{c}{NoPe-NeRF}              &  & \multicolumn{3}{c}{RoDynRF}              \\ 
\cline{4-6}\cline{8-10}\cline{12-14}\cline{16-18}
                  &                         &\vline& $\text{RPE}_t \downarrow$ & $\text{RPE}_r \downarrow$ & ATE$ \downarrow$ &  & $\text{RPE}_t \downarrow$ & $\text{RPE}_r \downarrow$ & ATE$ \downarrow$ &  & $\text{RPE}_t \downarrow$ & $\text{RPE}_r \downarrow$ & ATE$ \downarrow$   &  & $\text{RPE}_t \downarrow$ & $\text{RPE}_r \downarrow$ & ATE$ \downarrow$    \\ 
\hline
\multirow{4}{*}{\rotatebox[origin=c]{90}{NOTR}} & 016 && \best0.884 & \best0.225 &\best0.329   & & \sbest0.886 &\best0.225 & \sbest1.282  && \tbest1.859& 0.308& \tbest22.967 && 1.946  &\tbest 0.296  & 31.280  \\
& 021 && \sbest0.747 & \sbest0.574 &\best1.900   & & \best0.746 & \best0.560 & \sbest 2.578 && 1.658& 0.670&\tbest 19.516 && \tbest1.619  &\tbest 0.615  & 21.216  \\
& 022 && \best1.417 & \best0.146 &\best1.456   & & \sbest1.461 &\best0.146 &\sbest 7.388 && \tbest3.577& \tbest0.342&\tbest43.451 && 4.194  & 0.603 & 55.443  \\
& 025 && \best0.720 & \sbest0.289 &\best0.568   & &\sbest 0.739 &\best0.287 &\sbest1.356 && 1.758& 0.478 &\tbest10.201 && \tbest1.600  & \tbest0.374 & 26.494 \\
\hline
\multirow{3}{*}{\rotatebox[origin=c]{90}{\scriptsize{Pandaset}}} & 011 && \sbest2.176 &\sbest 0.182 &\best0.813   & & \best2.147 &\best0.178 & \sbest3.162  &&\tbest 2.629& 0.522& \tbest10.983 && 3.410  & \tbest0.363 &  11.170  \\
& 027 && \sbest1.625 & \best0.215 &\best0.259   & & \best1.618 &\sbest0.217 &\sbest0.888 &&\tbest 2.189 & \tbest0.282&  17.059 && 6.300 & 6.137  & \tbest7.792  \\
& 027 && \best1.475 & \best0.292 &\best0.159  & &\sbest1.478 &\sbest0.293 &\sbest 0.474 && \tbest1.836&\tbest 0.397&15.413 && 5.481 &7.626 & \tbest6.252 \\
\bottomrule
\end{tabularx}
}

\label{table:pose}
\end{table*}
\subsection{Comparison of Camera Poses Estimation}
\label{EXPERIMENTS:C}
 Firstly, to obtain the complete camera trajectory we freeze the NeRF model and optimize the camera poses for the test viewpoints. Afterward, we align the recovered camera trajectory with the ground truth using the Umeyama method (Sim(3) with 7 DoF) and evaluate its accuracy. The quantitative results of camera pose estimation are presented in TABLE \ref{table:pose}. NoPe-NeRF is not designed for dynamic scenes, while RoDynRF performs poorly in large-scale dynamic scenes such as urban driving due to its limited model capacity and inaccurate motion masks. Although LocalRF is not suitable for dynamic scenes, it can progressively optimize camera poses and NeRF, making it robust and accurate for long camera trajectories. However, due to its inability to effectively handle dynamic objects' interference, the recovered camera trajectories are less accurate than ours, especially in terms of global trajectory accuracy and consistency, where the gap is more noticeable. We visualize the camera trajectories in Fig. \ref{fig:pose} for qualitative comparison. Note that for a more detailed comparison, we set the coordinate systems of VDNeRF and LocalRF to the same range.

\subsection{Ablation Study}
\label{EXPERIMENTS:D}
\begin{figure}[ht]
    \centering
    \includegraphics[width=0.85\linewidth]{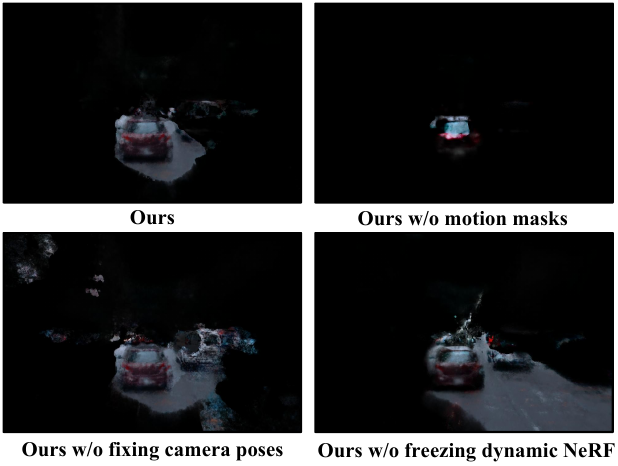}
    \caption{\textbf{Comparison of static and dynamic decomposition results of ablation studies.} The full VDNeRF achieves the best performance in static-dynamic decomposition and dynamic object rendering quality.}
    \label{fig:ablation}
\end{figure}

We conduct ablation studies around our training framework (motion masks, fixing the camera poses, freezing the dynamic NeRF). We analyze them in TABLE \ref{table:ablation}. We show the static-dynamic decomposition results of the ablation studies in Fig. \ref{fig:ablation}. Below we analyze in detail to further verify the effectiveness and reasonable of our training framework.

\begin{table}[ht]
\centering
\caption{\textbf{Ablation study results on the NOTR dataset.} We report PSNR, SSIM, and LPIPS of novel view synthesis, $\text{RPE}_t$, $\text{RPE}_r$ and ATE of the camera pose estimation. $\uparrow$: higher is better, $\downarrow$: lower is better.}
\setlength{\tabcolsep}{2pt}
\resizebox{\linewidth}{!}{
\begin{tabular}{lccccccc} 
\toprule
                      & \multicolumn{3}{c}{Novel View Synthesis}                               &  & \multicolumn{3}{c}{Camera Poses}                                               \\ 
\cline{2-4}\cline{6-8}
                      & PSNR$\uparrow$ & SSIM $\uparrow$ & LPIPS $\downarrow$ &  & $\text{RPE}_t\downarrow$ & $\text{RPE}_r\downarrow$ & ATE $\downarrow$  \\ 
\hline
full                  & \textbf{33.133} & \textbf{0.920}  & 0.186      &  & 0.884           & \textbf{0.225}           & \textbf{0.329}   \\
w/o motion masks & 32.738          & 0.917            &0.196               &  & \textbf{0.883}         & \textbf{0.225}                    & 0.662           \\
w/o fixing poses        & 33.115          & 0.919            &\textbf{0.185}              &  & 0.884                    & 0.226                    &  0.561            \\
w/o freezing $F_{\Theta}^{d}$   &  28.834          & 0.825           & 0.310      &  & 0.925                   & 0.518& 1.740            \\
\bottomrule
\end{tabular}
}
\label{table:ablation}
\end{table}

\textbf{Effect of motion masks.} With the help of motion masks, VDNeRF can exclude dynamic objects in the scene when jointly optimizing the camera poses and NeRF. This improves the robustness and accuracy of camera pose estimation, because, for dynamic objects, the motion of pixels on the image plane is caused by both the camera motion and the independent motion of the dynamic objects. Therefore, without motion masks, the global consistency and accuracy of the camera poses are somewhat affected, which is reflected in an increase of 0.333 in ATE. The reconstruction and decomposition quality of dynamic objects also worsens. In addition, we encourage objects contained in the motion mask to be more easily regarded as dynamic objects, which will also improve the effect of static-dynamic decomposition.

\textbf{Effect of fixing the camera poses.} Constantly optimizing the camera poses during the training process does not have a significant impact on novel view synthesis, but it does cause the accuracy of the recovered camera trajectory to deteriorate. The scene is reconstructed jointly by static NeRF and dynamic NeRF, achieving static-dynamic decomposition in a self-supervised manner. Once dynamic NeRF is activated and begins rendering dynamic objects, errors in the initial stage of decomposition can affect the rendering quality of the static background, leading to inaccuracies in camera pose optimization. Consequently, ATE increases by 0.329. Furthermore, continuous adjustments to camera poses, even minor ones, can lead to incorrect modeling of independent object motion. This is manifested in that some static pixels are misclassified as dynamic, and dynamic objects cannot be individually decoupled.

\textbf{Effect of freezing the dynamic NeRF.} When optimizing the camera poses, activating the dynamic NeRF $F_{\Theta}^{d}$ not only leads to inaccurate camera pose estimation but also causes poor image rendering quality. Because VDNeRF misinterprets the pixel motion caused by the camera movement as a dynamic object, preventing it from learning the correct scene representation and camera trajectory. As a result, for novel view synthesis, PSNR decreases by 4.299, while LPIPS increases by 0.124. Additionally, it tends to treat the entire motion trajectory of an object as a single dynamic for decoupling.

\subsection{Limitations}
\label{EXPERIMENTS:E}
Although VDNeRF can jointly optimize the camera pose and NeRF in dynamic urban scenes and achieve static-dynamic decomposition, it still has certain limitations. First, it may struggle with complex camera motion trajectories, such as rapid in-place 360-degree rotations. In such cases, the large angular displacement between adjacent images results in minimal overlap of observable regions, making feature matching unreliable and thereby reducing the continuity and consistency of the estimated camera trajectory. Secondly, its segmentation of dynamic objects lacks fine granularity, often leading to the over-segmentation of an entire region when multiple dynamic objects are in close proximity. Lastly, VDNeRF currently cannot meet the low-latency requirements of real-time systems such as robotic navigation. VDNeRF is better suited for offline reconstruction, simulation, or as a source of priors for real-time systems.
\section{Conclusions}
In this paper, we propose VDNeRF, which jointly optimizes camera poses and NeRF using only a set of images as input, without relying on ground-truth camera poses or depths. 
To achieve this, we carefully design two NeRF models, one for static background and another for dynamic objects, along with a tailored training framework. 
Through extensive quantitative and qualitative evaluation of mainstream urban driving datasets, VDNeRF demonstrates outstanding robustness and performance, outperforming state-of-the-art NeRF-based methods. In future work, we will incorporate continual learning capabilities \cite{cheng2024continual,cheng2025achieving} into our framework to enhance its adaptability across diverse urban scenarios.

\bibliographystyle{IEEEtran}
\bibliography{reference}










\begin{IEEEbiography}[{\includegraphics[width=1in,height=1.25in,clip,keepaspectratio]{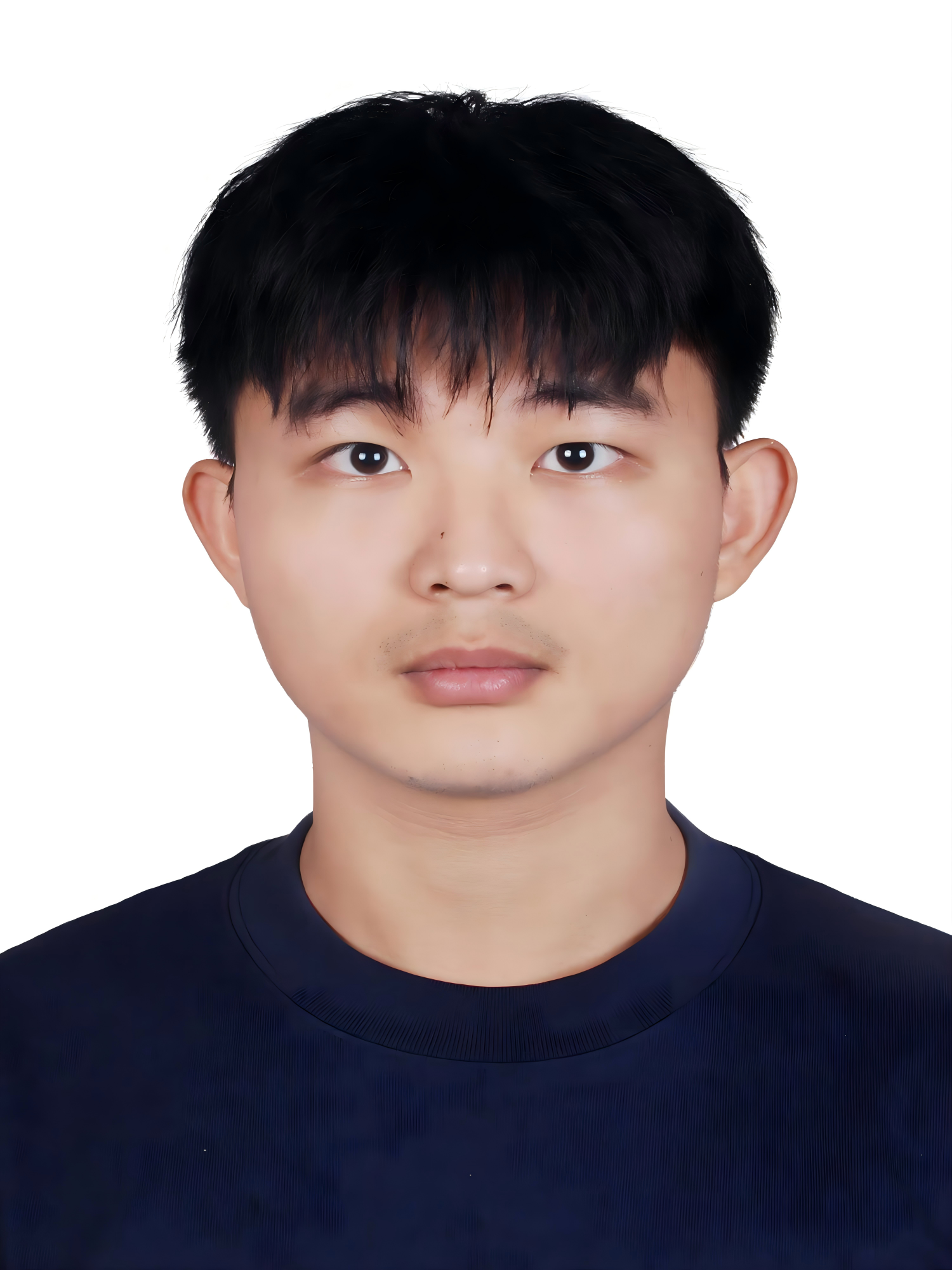}}]{Zhengyu Zou} received the B.E. degree from Hunan University, Changsha, China, in 2024. He is currently working toward the M.E. degree in the School of Automation, Northwestern Polytechnical University, Xi'an, China. His research interests include computer vision, novel view synthesis, 3D scene representation and reconstruction, and vision-based localization.
\end{IEEEbiography}

\begin{IEEEbiography}[{\includegraphics[width=1in,height=1.25in,clip,keepaspectratio]{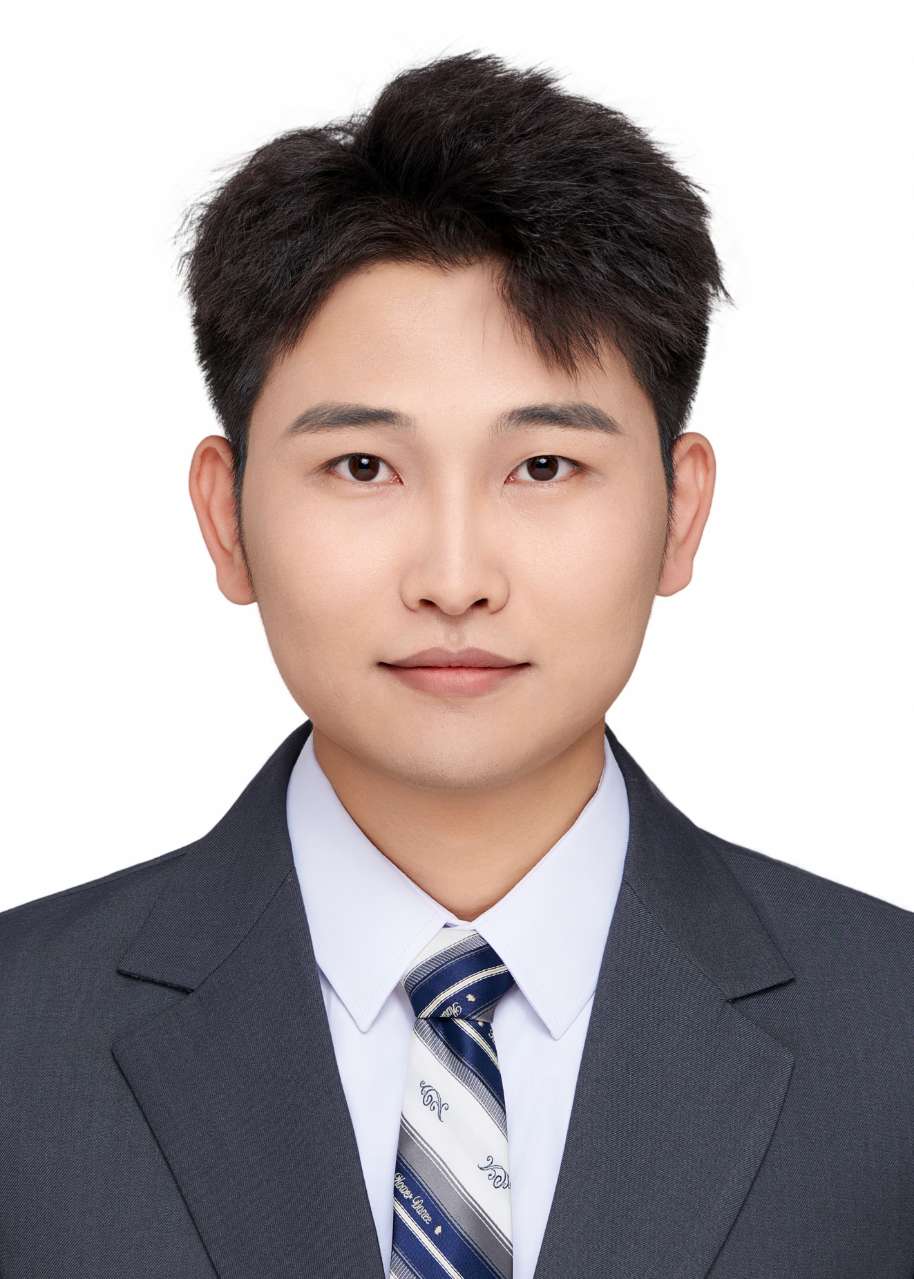}}]{Jingfeng Li}
received the B.E. degree from Southeast University, Nanjing, China, in 2022. He is currently pursuing the M.E. degree in the School of Automation, Northwestern Polytechnical University, Xi’an, China. His research interests including computer vision and pattern recognition, especially on visual localization and 3D scene reconstruction.
\end{IEEEbiography}

\begin{IEEEbiography}[{\includegraphics[width=1in,height=1.25in,clip,keepaspectratio]{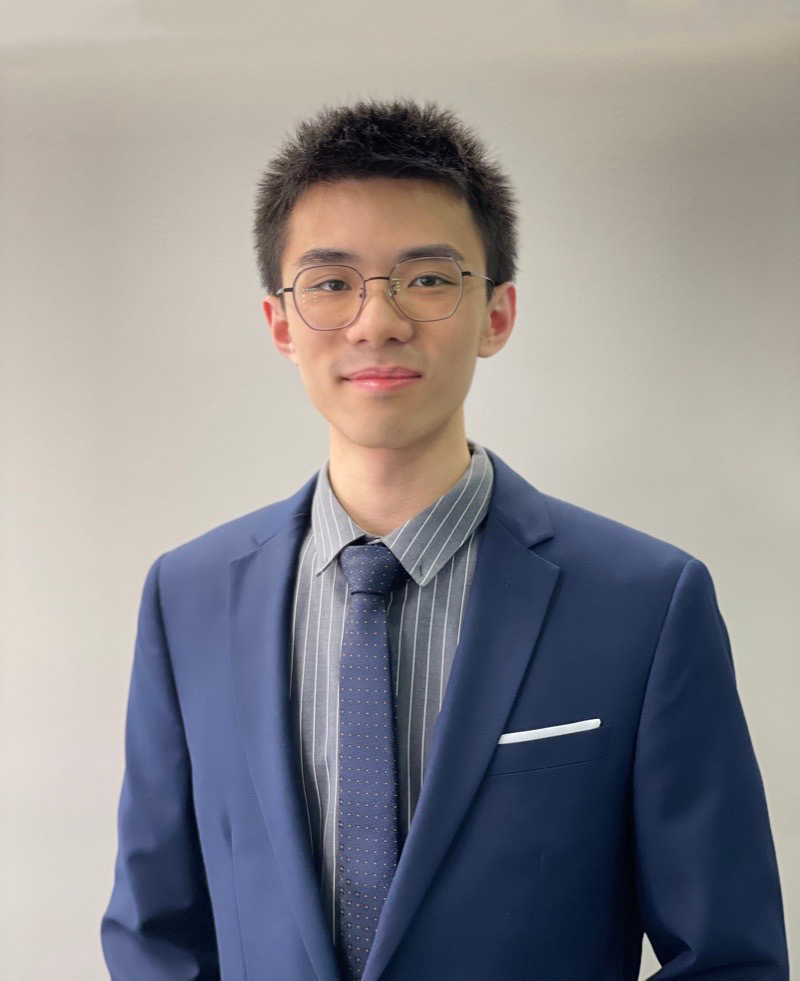}}]{Hao Li}
received the B.E. degree from Beijing University of Chemical Technology, Beijing, China, in 2022. He is currently working toward the Ph.D. degree in the School of Automation, Northwestern Polytechnical University, Xi'an, China. His research interests include computer vision and pattern recognition, especially on weakly supervised object segmentation, instance segmentation and 3D scene understanding.
\end{IEEEbiography}

\begin{IEEEbiography}[{\includegraphics[width=1in,height=1.25in,clip,keepaspectratio]{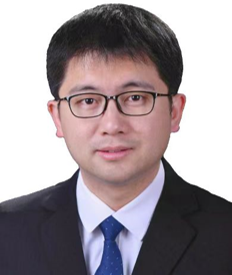}}]{Xiaolei Hou}
received the B.Eng. degree in automatic control from Harbin Engineering University, Harbin, China, in 2007, and the master degree in system and control and the doctor degree in robotics and computer vision from The Australian National University, Canberra, ACT, Australia, in 2010 and 2015, respectively. Between 2014 and 2016, he was a Postdoctoral Research Fellow with the Robotics and Computer Vision Group, Research School of Engineering, The Australian National University, where he is currently an Assistant Professor with the School of Automation, Northwestern Polytechnical University, Xi'an, China. His research interests include haptic teleoperation systems, aerial robotics, mobile robot navigation and path planning.
\end{IEEEbiography}

\begin{IEEEbiography}[{\includegraphics[width=1in,height=1.25in,clip,keepaspectratio]{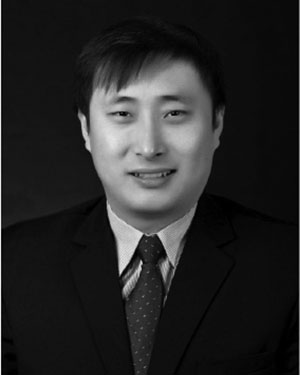}}]{Jinwen Hu}
(Member, IEEE) received the B.Eng. and M.Eng. degrees from Northwestern Polytechnical University (NPU), Xi’an, China, in 2005 and 2008, respectively, and the Ph.D. degree from Nanyang Technological University, Singapore. He is currently an Associate Professor with the School of Automation, NPU. His current research interests include multiagent systems, distributed control, unmanned vehicles, and information fusion.
\end{IEEEbiography}

\begin{IEEEbiography}[{\includegraphics[width=1in,height=1.25in,clip,keepaspectratio]{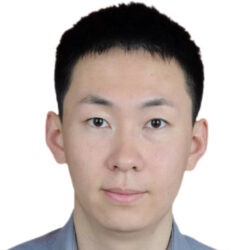}}]{Jingkun Chen}
Dr. Jingkun Chen is a postdoctoral researcher in the Department of Engineering Science at the University of Oxford, working with Professor Vicente Grau in the Oxford Biomedical Image Analysis (BioMedIA) group. His research focuses on medical artificial intelligence (AI), particularly in medical image analysis and multimodal data integration. He obtained his PhD from WMG, University of Warwick, where he developed machine learning methods such as multi-modal, weakly supervised, semi-supervised, and few-shot learning. These approaches are aimed at improving medical image interpretation and supporting clinical applications.

\end{IEEEbiography}

\begin{IEEEbiography}[{\includegraphics[width=1in,height=1.25in,clip,keepaspectratio]{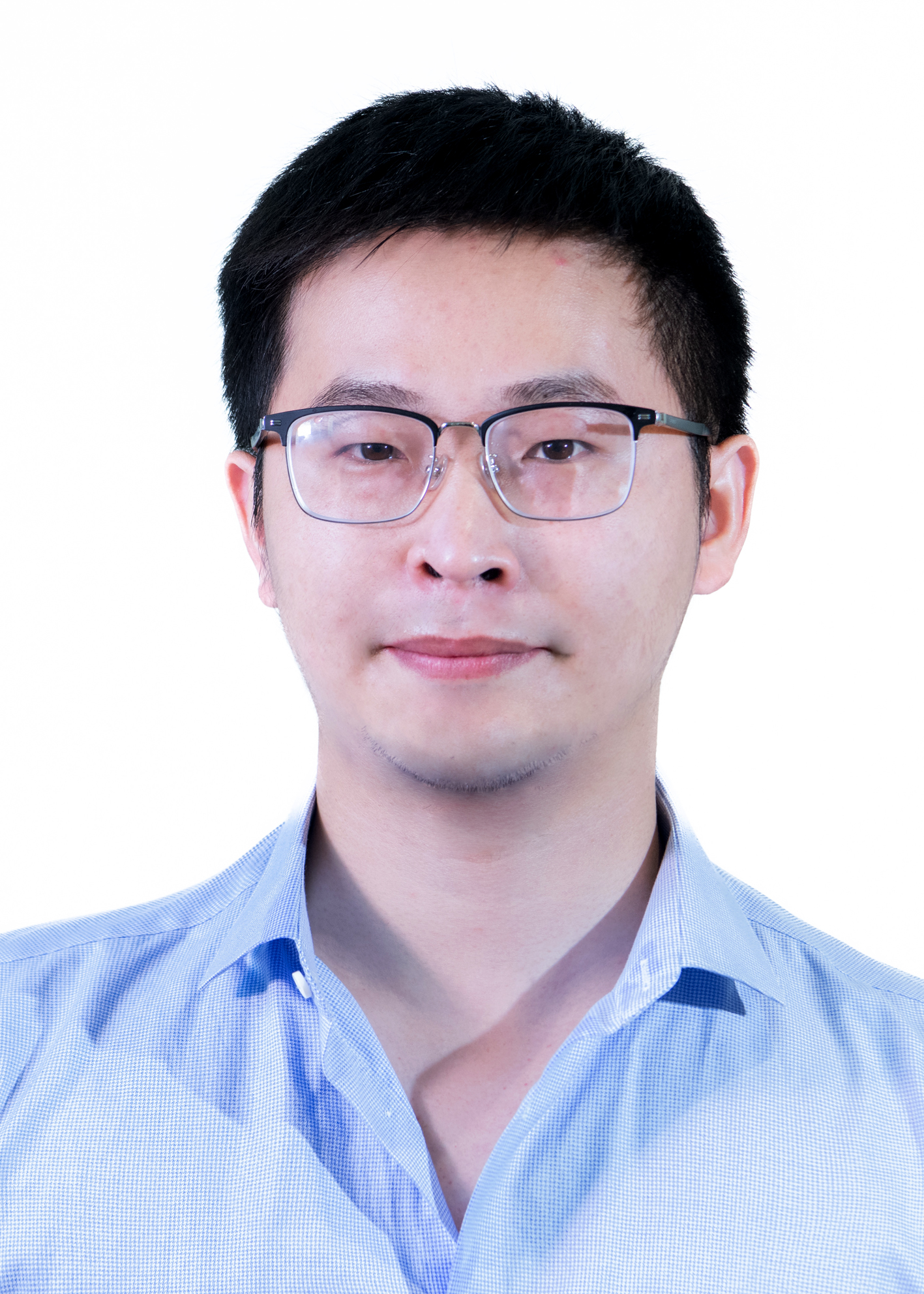}}]{Lechao Cheng}
received his Ph.D. degree from the Zhejiang University, HangZhou, China, in 2019. He is currently a Associate Professor in the School of Computer Science and Information Engineering, Hefei University of Technology. He has published over 40  top-tier conferences papers and journals, including CVPR, ICML, ECCV, ICCV. His research area centers around learning with defective data, model tuning, and media computing.
\end{IEEEbiography}

\begin{IEEEbiography}[{\includegraphics[width=1in,height=1.25in,clip,keepaspectratio]{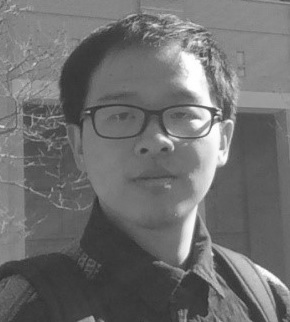}}]{Dingwen Zhang}
received his Ph.D. degree from the Northwestern Polytechnical University, Xi’an, China, in 2018. He is currently a Professor in the School of Automation, Northwestern Polytechnical University. From 2015 to 2017, he was a visiting scholar at the Robotic Institute, Carnegie Mellon University. His research interests include computer vision and multimedia processing, especially on saliency detection, video object segmentation, temporal action localization and weakly supervised learning.
\end{IEEEbiography}

\end{document}